# Design, manufacturing, and inverse dynamic modeling of soft parallel robots actuated by dielectric elastomer actuators

Jung-Che Chang, Xi Wang, Dragos Axinte, and Xin Dong

*Abstract*—**Soft parallel robots with their manipulation safety and low commercial cost show a promising future for delicate operations and safe human-robot interactions. However, promoting the use of electroactive polymers (EAPs) is still challenging due to the under-improving quality of the product and the dynamic modelling of the collaborations between multiple actuators. This article presents the design, fabrication, modelling and control of a parallel kinematics Delta robot actuated by dielectric elastomer actuators (DEAs). The trade-off between the actuation force and stroke is retaken by an angular stroke amplification mechanism, and the weight of the robot frame is reduced by utilizing 3D puzzling strip structures. A generic way of constructing a high-stability conductive paint on a silicon-based film has been achieved by laser scanning the DE-film and then sandwiching a conductive particle-based electrode with a paint which is mixed by the particles and photosensitive resin. Compared to the wildly used carbon grease, the fabricated electrode shows a higher consistency in its dynamic behaviour before and after the on-stand test. Finally, to predict the output force and inverse motion of the robot end effector, we constructed the inverse dynamic model by introducing an expanded Bergstrom-Boyce model to the constitutive behavior of the dielectric film. The experimental results show a prediction of robot output force with RSME of 12.4% when the end effector remains stationary, and a well-followed trajectory with less than RSME 2.5%.**

## I. INTRODUCTION

Soft or compliant behaviours of robots have recently become a new trend in the field of manipulation and gripping as the inconsistency and delicacy of both the target objects and application environments have become fundamental requirements (Majidi, 2014; Atia et al., 2022).

In the field of robots made of rigid materials, serial kinematic manipulators take the slender configuration and large working space as an advantage (Webster III and Jones, 2010; Dong et al., 2022). In contrast, manipulators with parallel kinematic chains mostly have their precision, payload, and high responding speed (Merlet, 2006; Camacho-Arreguin et al., 2022). Many industrial applications have benefited from these advantages such as electronic or food production (Iqbal et al., 2017; Krajcovic et al., 2013), biology and medical operation (Nakano et al., 2009; Li et al., 2018), and manufacturing (Axinte et al., 2018; Ma et al., 2019) etc. However, most manipulators with fully rigid components offer nearly infinite stiffness behaviour and are built with metallic materials. This leads to challenges when considering the safety issue of human-machine interaction and niche applications, such as Magnetic Resonance Imaging (MRI) guided surgery operations (Laski et al., 2015; Pile and Simaan, 2014).

To address the inherent softness of these materials, parallel kinematic mechanisms (PKMs) can be used as a complementary solution due to their stiffness and potential for high accuracy. However, the internal force coupling between the robot kinematics chains is a critical part of the analysis of the whole robot's behaviour. The recent research on compliant limbs (Borenstein, 1995; Park et al., 2012) and joints (Hongzhe and Shusheng, 2010; Zhao et al., 2012) has made the analysis and application of coupling behaviour caused by soft linkages (Black et al., 2017; Lindenroth et al., 2017), joints (Ma et al., 2020; Kozuka et al., 2013), and a combination of both (Moghadam et al., 2015; Garcia et al., 2020) possible. Different from the robot-independent models (Marchese and Rus, 2016; Wang and Simaan, 2019; Singh et al., 2018) consider the geometry of the soft linkages, robot-dependent models such as continuous Cosserat models (Renda et al., 2012; De Payrebrune and O'Reilly, 2017; Renda et al., 2018; Grazioso et al., 2019) and continuous Euler–Bernoulli beam models (Oliver-Butler et al., 2019; Fraś et al., 2014) take the material elasticity and the constraint force into consideration and provide the ability of the forward kinematic analysis for the soft PKMs. Recently, the absolute nodal coordinate formulation (ANCF) was utilised to analyse the soft behaviour of both soft actuators and the entire robot, which extends the modelling framework through spatial position gradients for a soft PKM constructed by three evenly deployed silicon pneumatic tube actuators (Huang et al., 2022). This fulfilled the need for real-time prediction of the forward and inverse kinematic and static behaviour of soft PKMs. However, to the authors' best knowledge, the inverse dynamic model between multiple soft actuators is still a grand challenge when considering the material hysteresis behaviour.

Some case studies already prove pneumatic and hydraulic soft actuators promising for human-machine interactions, such as rehabilitation (Wang and Xu, 2021) and extracorporeal ultrasound (Lindenroth et al., 2019). However, compared to the fluidic actuators, the compact design and less transmission loss of the electrical active actuators (e.g., dielectric elastomers actuators), make them more suitable for applications in environments that need a magnetic-free and safe operation such as MRI-guided surgery. To achieve this goal, three grand challenges need to be addressed.

Firstly, dielectric elastomers (DE) have been proven as a type of promising material, because of their high energy density (Pelrine et al., 2001; Shi et al., 2022) and large strain (Pelrine et al., 2000; Sriratanasak et al., 2022). Being deformed by charged compliant electrodes sandwiching the elastomer, dielectric elastomer actuators (DEAs) can generate a high push-to-weight ratio (Hau et al., 2018; Wang et al., 2024) and high response frequency (Zhu et al., 2010) under different configurations. The constitutive model of DE films has evolved from applying linear viscoelasticity models (Wissler and Mazza, 2007) to linearizing the material behaviour (Sarban et al., 2011), all until Ref. (Gu et al., 2017a) constructed a general modelling framework by parallelly expanding the nonequilibrium thermodynamics network which was initiated from Bergstrom–Boyce model (Bergström and Boyce, 1998). Although a satisfying prediction has been proved through forward validation (i.e. giving the voltage and predicting the behaviour). However, the application from the inverse perspective is still under discussion, especially when multiple DEAs collaborate to perform as parallel kinematics manipulators.

Secondly, the challenge of building DEAs with stable long-term behaviour, good manufacturing repeatability, and cost-effectiveness is one of the main barriers to further promoting the application of collaborations between multiple DEAs. Carbon grease is a commonly used material for the electrode as it holds good conductivity and good attachability to most of the materials. However, due to the high viscosity and hard-to-cure properties, the dehydration during the on-stand changes the conductivity and the stiffness of the electrode which deteriorates the behaviour of the DEA with time (Rosset et al., 2016). Alternative methods have shown satisfying results by methodologies such as directly stamping carbon nanotube (Duduta et al., 2016), Langmuir-Schaefer transferring a homogeneous layer of carbon nanotube to PDMS film (Ji et al., 2019; Ji et al., 2018), and mixing the carbon black or carbon nanotubes with double-component cured silicon, then pad-printing (Rosset et al., 2016) or spin coating (Xu et al., 2019) to the film. However, the former two methods require either expensive materials or complex fabrication techniques, while the latter approach requires a high temperature (~80ºC) which could lead to a challenge for the use of most of the plastic materials as external frames.

Thirdly, the limited stroke of DEAs is another challenge for the construction of a parallel robot when considering its workspace. Multilayer design is a commonly used approach to increase the output force or stroke (Youn et al., 2020; Guo et al., 2021) of the actuator by increasing the geometric dimension in a specific direction. However, the case for improving the output force reasonably dominates the research trend due to the large layer number when building multilayers along the actuating direction, as the thickness of a single layer is strictly limited to generate sufficient maxwell force during the actuator. Some specific configurations can also improve the stroke of the

DEA such as bistable mechanisms for a negative stiffness extension to the film (Baltes et al., 2022; Hodgins et al., 2011) and saddle shape DEA (Gu et al., 2018; Sriratanasak et al., 2022) as a large bending stroke. However, for the bistable DEA, the external mechanisms are bulky which decreases the integrity of the DEA. As for the bending DEA, the difficulties come from missing a proper point for the connection with external mechanisms, the three-dimensional deformation would further limit the construction of high-thickness multilayer film or stacking multiple DEAs for high output force. Overall, having a high-force DEA and adding an external trade-off between the stroke and force would be a wiser way to have a higher integration when involving DEAs for manipulator actuation.

To address the challenges, an inverse dynamics model for parallel kinematics mechanism driven by DEAs is developed, by using an expanded Bergstrom-Boyce constitutive model. Further, a plastic 3D puzzling strip constructed Delta PKM which is actuated by lozenge-shape DEAs (LSDEAs) has been built in this paper, and a four-bar-mechanism-based stroke amplifying lever has been designed for the trade-off between force and the stroke of the actuator. We increased the electrode quality by mixing the carbon black with photosensitive resins and further diluting in isopropyl alcohol (IPA) to form a thinner layer. The attachability between the film and the electrode is generically improved by laser scanning grids on the DE-film and pre-applying carbon black at the film surface. Comparisons have been made between the proposed paint and carbon grease in terms of dynamic behaviour changes over time after manufacturing. Last, the dynamic analysis has been done by introducing an expanded Bergstrom-Boyce model, and the output force prediction and inverse behaviour of constructed PKM have been validated under a payload test and trajectory test.

The main contribution of this paper has been summarized as follows:

1. We applied an inverse dynamic model to enable collaboration among multiple dielectric elastomer actuators (DEAs). Based on the Delta configuration and an expanded Bergstrom-Boyce model in reference (Gu et al., 2017a), our model considers the hyper-elasticity, hysteresis, and creeping of the film, as well as the geometrical constraints, external forces, and kinetic energy.

2. We proposed a generic process for sandwich-type electrode fabrication which involves laser scanning, solid conductive particles and photosensitive resin for an electrode with more stable behaviour (against a long period) and decent attachability between the electrode and the film. The comparison was reflected by the time-related performance of uniaxially loaded film samples using our proposed paint and carbon grease.

3. We constructed the robot with stroke amplifying levers based on 3D puzzling strip structures that significantly reduce the time and financial cost of building a lightweight robot.

The rest of the content is arranged as follows: section II introduces the design and fabrication of a DEA-based PKM. Both the concept of the stroke amplification mechanism (SAM) and the preparation of the proposed double-layered carbon paint electrode (DCPE) are recorded. An inverse model for the DEA-based PKM is presented in section III. The model allows an inverse control of the PKM and an output force prediction for the PKM end effector. The corresponding validation is recorded in section IV. This includes the comparison between the proposed DCPE and a general carbon grease electrode (CGE) of its ageing behaviour, the PKM trajectory and the payload test for end effector output force prediction. Finally, the discussion and conclusion are in section V.

## II. Design and Fabrication of the DEA-based PKM

In this section, a new design method of the DEA-based PKM is proposed with a new approach to DEA fabrications, enabling the long-term dynamic performance of the DEAs. Based on these proposed methods, a delta PKM was designed and fabricated, which will be later used for validating the inverse model of collaboration among multiple DEAs.

### 1. Concept of the DEA-based PKM

We propose a DEA-based PKM in which the plastic sheets with over-fitting 'hook' structures are the key for puzzle-like assembling. As shown in Fig. 1 (A), lozenge shape DEAs (LSDEAs) with four-bar linkages stroke amplification mechanisms (SAM) are the actuation units of the PKM. Each DEA consists of two biaxial tensioned DE films and the surface area increases when actuated (Fig. 1 (B)). Caused by the increment of the film area, the changed angle in the lozenge frame varied the output of the hard-connected SAM and finally led to a rotate-down motion to the corresponding PKM bicep. Two types of revolute hinges are used in the design. The snape hinge distributed in SAM and PKM joints holds a high off-axis stiffness which ensures the accuracy of the geometric movements, whereas the torsion joint distributed in LSDEA and PKM forearm offers a spring-back force and higher durability when receiving payload.

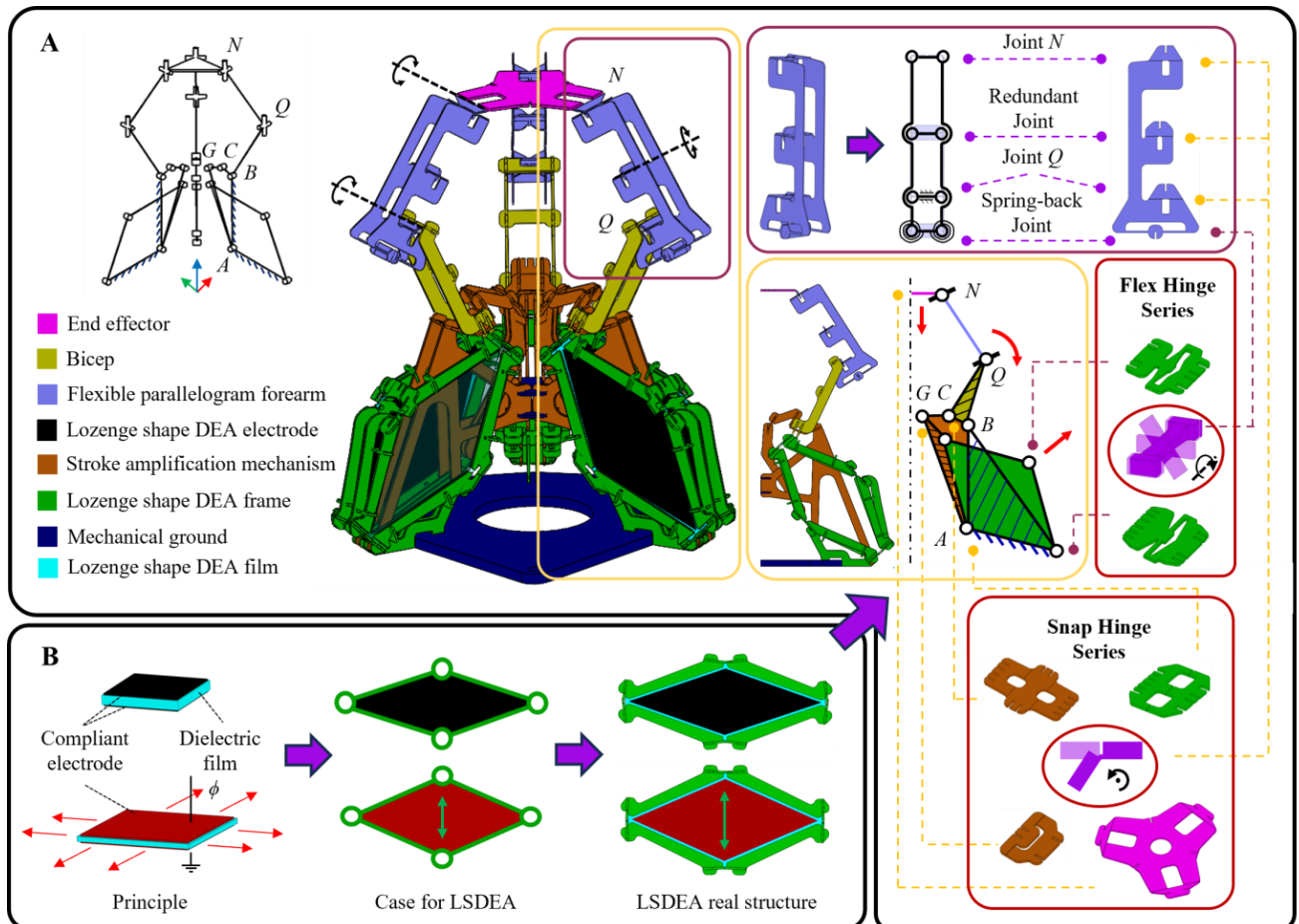

**Fig. 1** The principle and design of the DEA-based PKM. (A) Design of the DEA-based Delta PKM. The parallelogram forearm contains snape hinges and flexible hinges whose spring-back force forms an antagonist with the LSDEA. The DE-film is biaxially tensioned and fixed on lozenge-shaped rigid linkages (green). By actuating the lozenge shape DEA (LSDEA), the area of the LSDEA increases, which decreases the obtuse angle. A SAM (orange) is fixed on the LSDEA, and the output of the mechanism is fixed with a PKM bicep (yellow), which finally results as a clockwise motion of the bicep and a downward movement of the end effector. (B) Principle of the DEA. The actuator contains a dielectric elastomer (DE) film and two compliant electrodes clamping on the two sides of the film. Applying voltage to the electrodes, the generated maxwell force clamps the film and forms an extension on the rest two orthogonal directions of the DE-film.

Lozenge shape DEA are selected for a lighter weight, more stable long-term behaviour, and better repeatability of the actuator (Plante and Dubowsky, 2007; Moretti et al., 2020). Different from other actuators, the equivalent force on the two-diagonal direction of the lozenge shape frame is coupled by the linkage movement. Therefore, there is no necessity of using extra tension units (e.g. compressing spring). This eliminates the error caused by the large tolerance of the commercial springs, and so does the component creeping caused by the long-term compressing. Neglecting the spring-back force from the revolute joints, disusing the tension unit also simplifies the analysis of the unloaded DEA neutral state as the force balances between the two diagonal directions on the frame. This further enhances the quality control during the DEA preparation.

### 2. Preparation of the LSDEA

The challenges of constructing a stretchable compliant electrode mainly come from the need for high conductivity, low stiffness behaviour and the adhesion between the electrode and the film. As shown in Fig. 2 (A, E, F and G), we first double-side scanned the silicone film with a horizontal-and-vertical pattern which enlarges the attached amount of the carbon black powder during the painting. Then we mix carbon black with the elastic photosensitive resin (Elastic 50A, FORMLABS) for higher protectivity and strain. The stiffness from the electrode does not contribute to the theoretical DEA movement, which as an external payload reduces the overall DEA behavior. To minimize the negative contribution, we dilute the mixed resin (solute) with the evaporation solvent (isopropanol) to create a small thickness electrode by decreasing the viscosity of the uncured electrode (liquid) for the film formation. After painting the liquid onto the carbon black-covered film, the solvent inside the liquid starts to evaporate and leaves a thin solute layer on the film which is to be cured under the UV light and finally forms a thin protective compliant electrode.

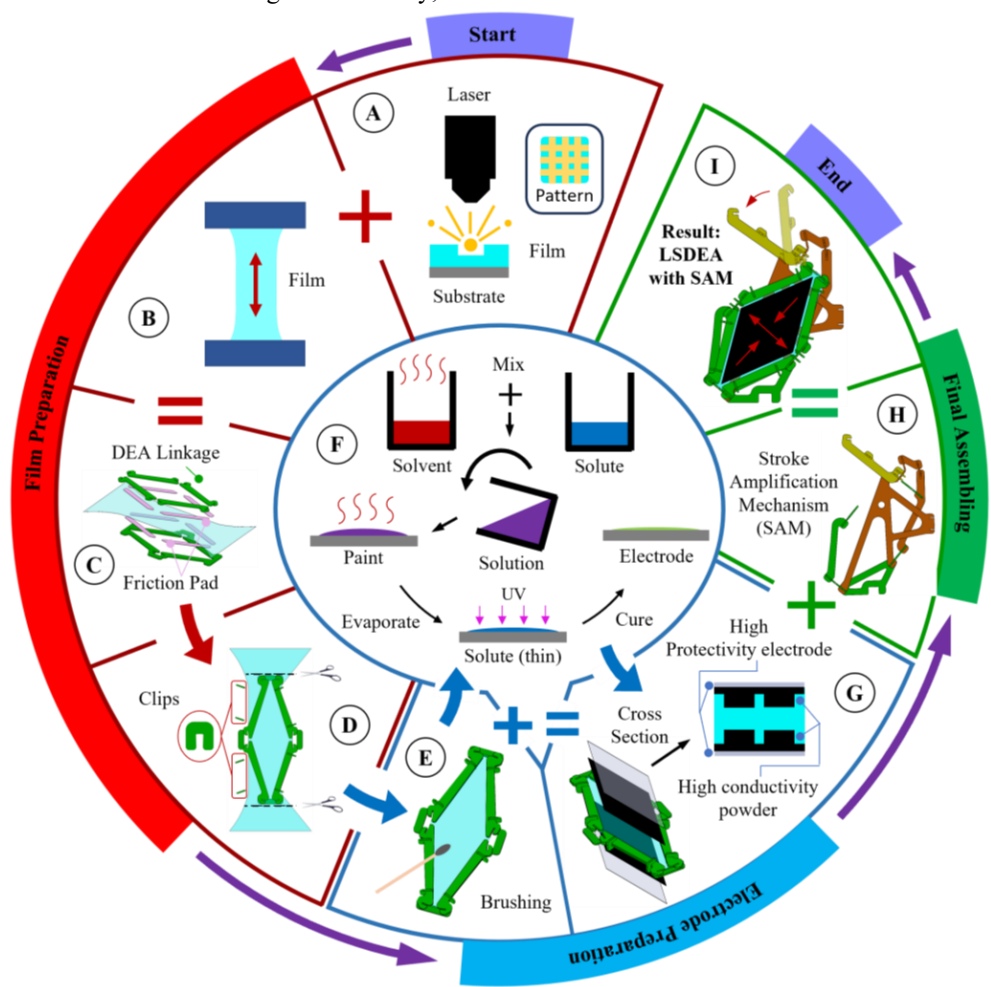

**Fig. 2** Fabrication process of an LSDEA with SAM. During the film fixing process (red), the DE-film was firstly double sides scanned with horizontal-and-vertical pattern (A), then uniaxially tensioned (B) and sandwiched by plastic frames with friction pads (C), and finally clamped by 'C' shape clips and trimmed from the pre-tension mechanism (D). During the Electrode preparation process (blue), both sides of the film were firstly brushed with carbon black powder (E) as the main conductive layer, then overpainted with the compliant electrode which is made by a mixture of photosensitive resin, solvent, and conductive powder, and finally cured under UV light after evaporation (F). This forms a DEA with double-layer electrodes which have a decent stretching ratio and good self-protectability (G). With two films ready, the final assembling process (green) was done by hooking the strips (linkages) to construct the SAM with bicep and hinges (H). Finally, a single LSDEA with SAM was prepared by releasing the lock of the DEA frames and hooking them to the two sides of the SAM, and the film rebalanced into a biaxially tensioned state (I).

The full construction of an LSDEA with SAM is shown in Fig. 2, summarizing into three phases: Film preparation (red), electrode preparation (blue) and final assembling (green). Aiming for faster DEA preparation, the use of non-instant adhesion should be avoided. During the film preparation phase, we uniaxially pre-tensioned the scanned DE-film and sandwiched it with friction pads (double-sided tape) and frames. Any physical damage inside the film can lead to film fracture due to the stress concentration during the pre-tension. Therefore, when fixing the DE-film to the frame, 'C' shape clips are utilized instead of screws to avoid any possible film damage. Following the aforementioned electrode preparation steps, we can finally mount the readied DEAs onto the assembled mechanism and become a desired lozenge shape DEA with a stroke amplification mechanism.

As a summary, this section details the principle and design of the DEA-based PKM. The overall structure is based on the laser cut 3D puzzling strip structures and the preparation of the actuator involves a generic electrode preparation method for more stable long-term dynamic behaviour and better attachability for the commercial stretchable materials to the silicon-based film.

## III. Inverse Modelling of the DEA-based PKM

To enable the collaboration between multiple DEAs, we derived a new virtual work-based approach for the proposed parallel kinematic mechanism to predict its dynamic behavior and the output force of the end effector. The virtual-work-based principle (framework) of the model and the correlation between the design parameters and the theoretical parameters within an unloaded LSDEA are shown in Sec. III 1. Sec. III 2 shows a geometrical model to address the conversion from the end effector position to the two directional strains of the DE-film on each LSDEA (for inverse analysis). Finally, the modelling of the DE-film constitutive behaviour is introduced in Sec III 3., in which the hyper-elasticity behaviour is modelled by the Gent model (Gu et al., 2017b), and the viscoelasticity is modelled by an expanded Bergstrom-Boyce model(Gu et al., 2017a).

### 1. Principle and design parameters

Aiming to predict the applied time-dependent voltage $\phi(t)$ under a specified time-dependent end-effector position (i.e. path) tensor $\mathbf{u} = [u_x(t), u_y(t), u_z(t)]$ (in x, y and z axis under global coordinate 0), by neglecting the weight of DEA and SAM, we constructed an inverse dynamic model which works under medium and low speed through the balance of external and internal virtual work as below:

$$\delta W_{\mathrm{PL}}\left(\mathbf{F}, \mathbf{u}\right) + \delta W_{\mathrm{INRT}}\left(\mathbf{u}\right) + \delta W_{\mathrm{VOLT}}\left(\phi, \mathbf{u}\right) = \delta W_{\mathrm{ELEC}}\left(\phi, \mathbf{u}\right) + \delta W_{\mathrm{ELAS}}\left(\mathbf{u}\right) \qquad (1)$$

where $\delta W_{\mathrm{PL}}$ is the virtual work from the external payload, which contains the end effector payload, the mass of both the end effector and forearm, and forearm spring-back force (collectively referred to as a time-dependent spatial force tensor $\mathbf{F} = [F_x(t), F_y(t), F_z(t)]$); $\delta W_{\mathrm{INRT}}$ is the virtual work from the inertial of the end effector mass, forearm mass, and external payload; $\delta W_{\mathrm{VOLT}}$ arises from the external voltage; $\delta W_{\mathrm{ELEC}}$ arises from the electric field; $\delta W_{\mathrm{ELAS}}$ is contributed by the in-plane deformation of the DE-film.

Further expanding the mentioned virtual works by the physical properties, the virtual works under a specified instant (i.e. time) are expressed as below:

$$\begin{cases} \delta W_{\mathrm{PL}} = \mathbf{F} \cdot \delta \mathbf{u} \\ \delta W_{\mathrm{INRT}} = m\ddot{\mathbf{u}} \cdot \delta \mathbf{u} \\ \delta W_{\mathrm{VOLT}} = \phi \delta Q \\ \delta W_{\mathrm{ELEC}} = \int \dfrac{D}{\varepsilon} \delta D dV \\ \delta W_{\mathrm{ELAS}} = \int \boldsymbol{\sigma} \cdot \delta \boldsymbol{\lambda} dV \end{cases} \qquad (2)$$

where $\delta \mathbf{u} = [\delta u_x, \delta u_y, \delta u_z]$ and $\ddot{\mathbf{u}} = [\ddot{u}_x, \ddot{u}_y, \ddot{u}_z]$ are the virtual displacement tensor and real acceleration tensor of the end effector, which can be obtained from the first and second derivative of $\mathbf{u}$; $m$ is the total mass of the mounted object plus the end effector; $\delta Q$ and $Q$ are the virtual and real increment of the electrical charge, where $Q = DS$; $S = \lambda_1 \lambda_2 \lambda_2 L_2$ is the area of the deformed DE film; $\delta D$ and $D = \varepsilon \phi \lambda_1 \lambda_2 / L_3$ are the virtual and true electric displacement in the elastomer; $\varepsilon$ is the permittivity of the DE material; $V$ is the volume of the DE-film; $\delta \boldsymbol{\lambda} = [\delta \lambda_1, \delta \lambda_2]$ is the in-plane virtual strain tensor of the DE-film along the two orthogonal directions; $\boldsymbol{\sigma} = [\sigma_1, \sigma_2]$ is the in-plane real stress tensor of the DE-film.

This subsection mainly addresses the mindset of the constructed model. For the application, a kinematic model of the mechanism and the constitutive behavior of the DE material should be induced. Then the correlation between $\delta \lambda_{k1}$, $\delta \lambda_{k2}$ and the virtual displacement under world coordinate origin $\delta \mathbf{u}^0$ could be constructed by the derivative of the inverse kinematic behaviour, which can be expressed as:

$$\delta \lambda_{ki} = \dfrac{\partial K_{ki}^{\mathrm{DEA}}}{\partial \theta_k^{\mathrm{DEA}}} \nabla K_{k\mathrm{DEA}}^{\mathrm{inv}} \cdot \mathbf{P}_k^0 \mathbf{P}_k^0 \left(\delta \mathbf{u}^0\right)^T \qquad (3)$$

where $K_{ki}^{\mathrm{DEA}}$ is the kinematic model in kinematic chain $k = 1, 2, 3$ from the DEA angle $\theta_k^{\mathrm{DEA}}$ to the corresponding film strain in direction $i$ (shown in Fig. 4); $K_{k\mathrm{DEA}}^{\mathrm{inv}}$ is the inverse kinematic model of the PKM from the tip of the forearm $\mathbf{N}_k^j$ to the DEA angle $\theta_k^{\mathrm{DEA}}$ on kinematic chain $k$ (shown in Fig. 4 ) and $\mathbf{P}_k^0$ is the projection matrix for the conversion from $\delta \mathbf{u}^0$ to $\delta \mathbf{u}_k^0$ which is contributed by kinematic chain $k$.

As shown in Fig. 3, the biaxial-tensioned film is completed first by an uniaxially pre-tension on the primary axis, then once the fixed frame and film were cut off from the pre-tensioning tool, the lock on the frame unlocked the position and a tensioning on the secondary axis is automatically achieved by a lateral force balancing. Therefore, $K_{ki}^{\mathrm{DEA}}$ could be geometrically expressed as:

$$\begin{cases} K_{k1}^{\text{DEA}} = \dfrac{2l_{\text{DEA}}}{L_1} \sin\left(\dfrac{\theta_k^{\text{DEA}}}{2}\right) \\[3mm] K_{k2}^{\text{DEA}} = \dfrac{2l_{\text{DEA}}}{L_2} \cos\left(\dfrac{\theta_k^{\text{DEA}}}{2}\right) \end{cases} \tag{4}$$

where $L_1$ and $L_2$ are the in-plane two-dimensional neutral length of the DE film.

For the case of predicting end effector output force $\mathbf{F}_u$ in the proposed DEA-based PKM, the force on three tips of the forearms is separately analyzed and finally synthesized to the end effector, which can be expressed as:

$$\mathbf{F}_u = \sum_{k=1}^{3} \mathbf{F}_k \tag{5}$$

where $\mathbf{F}_k$ ($k = 1,2,3$) is the force at the tip of the forearm in kinematic chain $k$. To predict its composition force which is contributed by the corresponding DE-film, we inherit the virtual-work-based framework by substituting the input into $\mathbf{F}_k$, $\phi$ and $\mathbf{u}$, which are expressed as below:

$$\delta W_{\text{PL}}\left(\mathbf{F}_k, \mathbf{u}\right) + \delta W_{\text{INRT}}\left(\mathbf{u}\right) + \delta W_{\text{VOLT}}\left(\phi, \mathbf{u}\right) = \delta W_{\text{ELEC}}\left(\phi, \mathbf{u}\right) + \delta W_{\text{ELAS}}\left(\mathbf{u}\right) \tag{6}$$

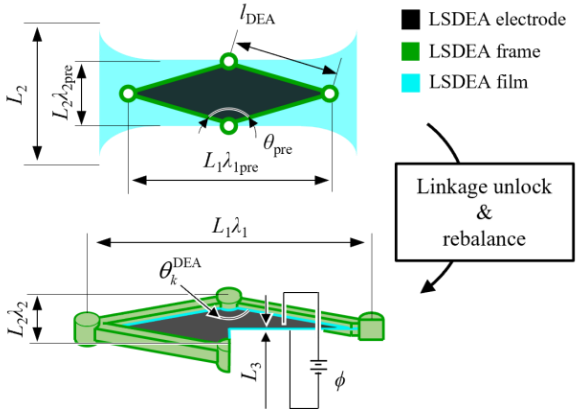

**Fig. 3** Condition of the DE-film before and after the two-phased biaxial tensioning. The DE-film is uniaxially pretensioned and fixed on a lozenge-shaped mechanism (frame) where the joints of the mechanism are locked. Then, the film is trimmed down from the pretension tool, and the frame joint is unlocked, which enables the rebalancing in the two directions, and finally forms a stable unactuated state.

Additionally, considering the non-intuitive correlation between the prefined DE film parameters and the dimensions of the non-tensioned DE-film, below we constructed their correlation by applying the Gent model to predict the hyper-elastic behavior of the film. Hence, the relation between the two strains after the uniaxial pre-tensioning $\lambda_{1\text{pre}}$ and $\lambda_{2\text{pre}}$ can be derived by giving a free boundary constraint to the film's secondary axis:

$$\frac{\mu_1\left(\lambda_{2\text{pre}}^2 - \lambda_{1\text{pre}}^{-2}\lambda_{2\text{pre}}^{-2}\right)}{1 - \left(\lambda_{1\text{pre}}^2 + \lambda_{2\text{pre}}^2 + \lambda_{1\text{pre}}^{-2}\lambda_{2\text{pre}}^{-2} - 3\right)/J} = 0 \tag{7}$$

where $J$ represents a constant which relates to the maximum tension limit of the film, and $\mu_1$ is the shear-strain modulus of the hyper-elastic spring. Therefore, according to Eqs. (4, 7), the actuator-level design parameters: preset angle of the lozenge shape linkage $\theta_{\text{pre}}$, the linkage length $l_{\text{DEA}}$, film pre-tension strain $\lambda_{1\text{pre}}$, $\lambda_{2\text{pre}}$, and $L_1$, $L_2$ follows the below relation:

$$\begin{cases} \lambda_{2\text{pre}} = \lambda_{1\text{pre}}^{-0.5} \\[2mm] L_1 = 2l_{\text{DEA}}\lambda_{1\text{pre}}^{-1}\sin\left(\dfrac{\theta_{\text{pre}}}{2}\right) \\[3mm] L_2 = 2l_{\text{DEA}}\lambda_{1\text{pre}}^{0.5}\cos\left(\dfrac{\theta_{\text{pre}}}{2}\right) \end{cases} \tag{8}$$

As summary, we constructed a generic framework for the collaboration of the multiple DEAs. Specifying the proposed prototype, the correlations between the designed parameter and the parameter for the prototype preparation are explained.

### 2. Kinematics of the DEA-based PKM frame

We constructed the kinematics of the PKM frame through a geometric approach. Each chain of the PKM holds the same design parameter. Therefore, by projecting the position of the end effector from the world coordinate to each of the local (joint) coordinates, the inverse analysis is done by processing the same analysis three times with different inputs (i.e. end effector position under respective joint coordinate). Fig. 4 (b) shows a loop for single actuator analysis. This subsection derives the conversion from the task space to the configuration space. The full conversion from the configuration space to the actuator space would need the participation of the DE-film constitutive behaviour, which is introduced in Sec. III 3.

Joint $N_k$ showing in Fig. 4 (a) is rigidly fixed to the end effector. Therefore, the position of joint $N_k$ under the local coordinate system can be described by a series of rigid-body transformations by the given end effector position, which could be expressed as:

$$\mathbf{N}_k^j = {}^0\mathbf{M}_j \mathbf{M}_{kN}^U \left(\mathbf{u}^0\right)^T \tag{9}$$

where $\mathbf{N}_k^j = \left[N_{kx}^j, N_{ky}^j, N_{kz}^j\right]$ is the Cartesian position of the joint $N_k$ on kinematic chain $k = 1, 2, 3$ based on the local coordinate system in actuator $j = 1, 2, 3$; $\mathbf{u}^0 = \left[u_x, u_y, u_z\right]$ is the Cartesian position of the end effector based on the world coordinate system 0; ${}^0\mathbf{M}_j$ is the transformation matrix to transform the coordinate from 0 (world) to $j$ (local), which is correlated to the geometric parameters shown in Fig. 4 (C); $\mathbf{M}_{kN}^U$ is the prismatic transformation matrix from $\mathbf{u}^0$ to $\mathbf{N}_k^0$ on the kinematic chain $k$, which is correlated to the geometric parameter. The following analysis will be focused on a single kinematic chain.

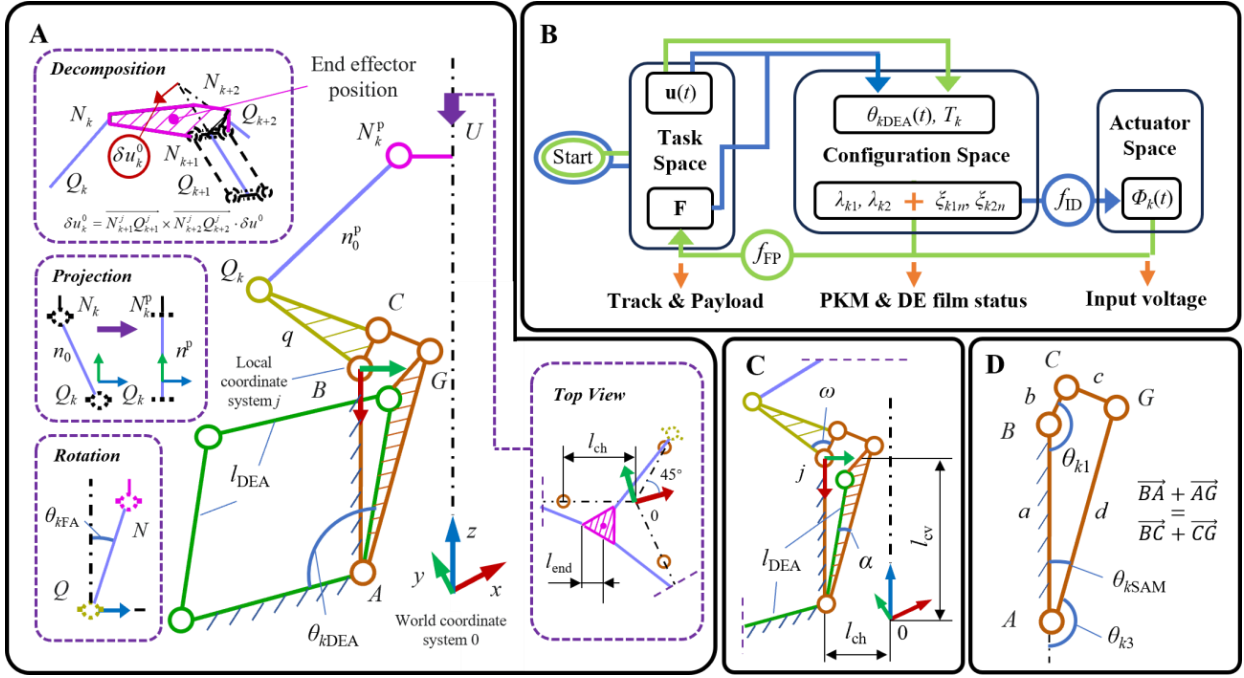

**Fig. 4** Analysis of the DEA-based PKM frame. (A) Schematic of a single chain in the DEA-based PKM frame. The virtual increment of the end effector can be decomposed into three directions which is correspondingly contributed by the three chains. The direction of the virtual work in each chain is obtained by taking the norm of the two forearms in the rest of the two PKM chains. While deriving the kinematic behaviour of the PKM frame, the forearm is projected to the datum of the analyzing actuator. To analyze the spring-back force of the forearms, the rotational angle is obtained through the angle between the forearm and the z-axis of the corresponding local coordinate. (B) Algorithm of the flow for dynamic analysis. $f_{ID}$ stands for the function of the inverse dynamic where the inputs are the time-dependent path and the time-dependent payload, and the output is the time-dependent voltage applied to each of the actuators. $F_{FP}$ stands for the function of end effector output force prediction where the inputs are the time-dependent track and the time-dependent voltage applied to each of the actuators, and the output is the instant output force of the end effector. (C) Schematic of the design parameters in the DEA-based PKM frame. (D) Schematic for the analysis of the stroke amplification mechanism. The geometric formula could be formed by the close loop of the four linkages.

We flattened the subsequent analysis into in-plane motion by projecting the real position of joint $N_k$ into the local $xy$ plane. Therefore, the projected linkage length $e^p$ in actuator $j$ and the output of the corresponding SAM $\theta_{k1}$ can be geometrically expressed as:

$$\begin{cases} \theta_{k1} = \pi - \arctan\left(\dfrac{N_y^j}{-N_x^j}\right) + \arccos\left(\dfrac{q^2 + \left(l_{BN}^p\right)^2 - \left(n_0^p\right)^2}{2q l_{BN}^p}\right) - \omega \\ n^p = \sqrt{\left(n_0\right)^2 - \left(N_{kz}^j\right)^2} \\ l_{BN}^p = \sqrt{\left(N_{kx}^j\right)^2 + \left(N_{ky}^j\right)^2} \end{cases} \quad (10)$$

where $q = l_{BQ}$ is the geometrical length of the PKM bicep.

The amplification ratio of SAM depends on the length ratio of its four linkages. Defining the length of the four linkages respectively as $a$, $b$, $c$, $d$ (Fig. 4 (c)), for the inverse analysis, the correlation between $\theta_{k1}$ and the input of the SAM $\theta_{kSAM}$ could be geometrically expressed as below:

$$\begin{cases} \theta_{kSAM} = 180° - 2\tan^{-1}\left(\dfrac{-G - \sqrt{G^2 - 4HR}}{2H}\right) \\ G = -4bd \sin \theta_{k1} \\ H = a^2 + b^2 + d^2 - c^2 - 2ab \cos \theta_{k1} - 2ad + 2bd \cos \theta_{k1} \\ R = a^2 + b^2 + d^2 - c^2 - 2ab \cos \theta_{k1} + 2ad - 2bd \cos \theta_{k1} \end{cases} \quad (11)$$

where $G$, $H$ and $R$ are the algebraic intermediates.

The decomposition of $\delta \mathbf{u}^j$ to $\delta u_k^j$, the virtual displacement component contributed by kinematic chain $k$ is based on the geometrical constraint of the robot's Delta configuration. Defining the forearm direction on the rest of the two kinematic chains as $\mathbf{n}_{k+1}^j = \mathbf{Q}_{k+1}^j - \mathbf{N}_{k+1}^j$, $\mathbf{n}_{k+2}^j = \mathbf{Q}_{k+2}^j - \mathbf{N}_{k+2}^j$, the decomposition could be mathematically understood as a vector projecting to $\mathbf{P}_k^j$ the normal vector of a plane formed by $\mathbf{n}_{k+1}^j$ and $\mathbf{n}_{k+2}^j$, where the normal vector is in a modular of one. This can be expressed as:

$$\begin{cases} \delta u_k^j = \mathbf{P}_k^j \cdot {}^0\mathbf{M}_j \delta \mathbf{u}^0 \\ \mathbf{P}_k^j = \dfrac{\mathbf{n}_{k+2}^j \times \mathbf{n}_{k+1}^j}{\left|\mathbf{n}_{k+2}^j \times \mathbf{n}_{k+1}^j\right|} \end{cases} \quad (12)$$

The spring-back force in each of the forearm is treated as a part of the external payload, which the virtual work $\delta W_{kFA}$ can be expressed as:

$$\delta W_{kFA} = \left(\mathbf{P}_k^j \cdot \mathbf{F}_{kORTH}^j\right) \mathbf{P}_k^j \cdot \delta \mathbf{u}^j \quad (13)$$

where $\mathbf{F}_{kORTH}^j$ describes the force along the orthogonal direction of the forearm, which can be expressed by the forearm spring-back force (torque) $T(\theta_{kFA})$ as:

$$\mathbf{F}_{k\mathrm{ORTH}}^{j} = n_0 T \left( \frac{\mathbf{n}_k^j \times \mathbf{x}^j}{\left| \mathbf{n}_k^j \times \mathbf{x}^j \right|} \right) \tag{14}$$

where $\mathbf{x}^j = [1, 0, 0]^T$ is the unit vector under the local coordinate $j$; $\mathbf{n}_k^j = \mathbf{Q}_k^j - \mathbf{N}_k^{j}$ is the direction of the forearm in chain $k$ under its local coordinate $j$. And the position of the bicep tip $\mathbf{Q}_k = [Q_{kx}, Q_{ky}]$ can be expressed as:

$$\mathbf{Q}_k = q \boldsymbol{\theta}_{k1} \tag{15}$$

where we define $\boldsymbol{\theta}_{k1} = [\cos\ \theta_{k1}, \sin\theta_{k1}, 0]$ as the factor matrix projecting to the end position while an object (here refers to the bicep $Q_k B$) rotates with an angle of $\theta_{k1}$. Decomposing the total force $\mathbf{F}_k^j$ on joint $N_k$, we have the below relationship:

$$\mathbf{F}_k^j = \mathbf{F}_{k\mathrm{PARA}}^j + \mathbf{F}_{k\mathrm{ORTH}}^j \tag{16}$$

where $\mathbf{F}_{k\mathrm{PARA}}^j$ is the force along the forearm, which is contributed by $\boldsymbol{\sigma}_k(t)$.

The correlation between the end effector position and $\theta_{k\mathrm{FA}}$ the rotation angle of the parallel forearm can be expressed as:

$$\theta_{k\mathrm{FA}} = \arccos\left( \frac{\mathbf{z}^j \cdot \mathbf{n}_k^j}{n_0} \right) - \frac{\pi}{2} \tag{17}$$

where $\mathbf{z}^j = [0, 0, 1]$ is the unit vector under the local coordinate $j$, and $n_0$ is the length of the PKM forearm.

Therefore, by combining Eqs (9, 10, 15), $K_{Qk}^{\mathrm{inv}}$ the correlation between $\mathbf{N}_k^j$ and $\theta_{k1}$ can be obtained. Further, by combining Eqs (4, 9-11, 17), we can finally get the inverse kinematics of the PKM frame $K_{k\mathrm{DEA}}^{\mathrm{inv}}$.

### 3. Constitutive behavior of the DE-film

As shown in Fig. 5 , we introduced a seventh-order expanded Bergstrom-Boyce model (Gu et al., 2017a) to predict the hysteresis behaviour of the DE-film. For the film viscoelastic and creeping behaviour, we parallelly expanded the viscoelastic sub-chain in Bergstrom-Boyce model into six chains, and $\mu_n$ (where $n = 1, 2..., 7$) is the shear-strain modulus of the springs. The dashpots are assumed to behave as Newtonian fluid with viscosity $\eta_n$ (where $n = 2, 3..., 7$), hence $T_{n-1}$ (where $n = 2, 3..., 7$) the relaxation time in each viscoelastic sub-chains would be the division between the dashpot viscosity and the spring shear modulus, expressed as:

$$T_{n-1} = \eta_n / \mu_n \tag{18}$$

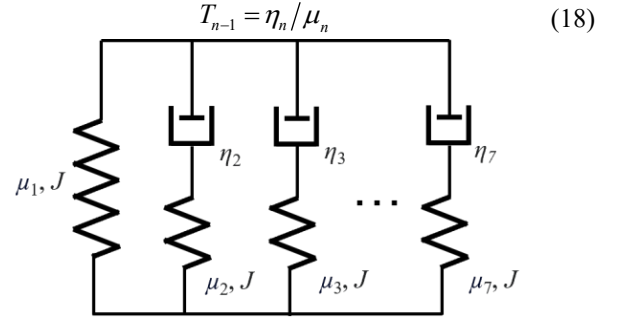

**Fig. 5** Expended Bergstrom-Boyce model for the nonlinear viscoelastic material constitutive behaviour.

Defining the response of the damper on chain $n$ as $\xi_n$ and the strain of the corresponding spring as $\lambda_n^e$, the total strain $\lambda_i$ at direction $i = 1, 2$ could be expressed as:

$$\lambda_i = \lambda_{ni}^e \xi_{ni} \tag{19}$$

And the coupled dynamic behaviour between $\xi_{n1}$ and $\xi_{n2}$ can be expressed as:

$$\begin{cases} \dfrac{1}{\xi_{n1}} \dfrac{d\xi_{n1}}{dt} = \dfrac{1}{3\eta_n} \left( \dfrac{\mu_n \left( \lambda_1^2 \xi_{n1}^{-2} - \lambda_1^{-2} \lambda_2^{-2} \xi_{n1}^2 \xi_{n2}^2 \right)}{1 - \left( \lambda_1^2 \xi_{n1}^2 + \lambda_2^2 \xi_{n2}^2 + \lambda_1^{-2} \lambda_2^{-2} \xi_{n1}^2 \xi_{n2}^2 - 3 \right) / J} - \dfrac{\mu_n \left( \lambda_2^2 \xi_{n2}^{-2} - \lambda_1^{-2} \lambda_2^{-2} \xi_{n1}^2 \xi_{n2}^2 \right) / 2}{1 - \left( \lambda_1^2 \xi_{n1}^2 + \lambda_2^2 \xi_{n2}^2 + \lambda_1^{-2} \lambda_2^{-2} \xi_{n1}^2 \xi_{n2}^2 - 3 \right) / J} \right) \\ \dfrac{1}{\xi_{n2}} \dfrac{d\xi_{n2}}{dt} = \dfrac{1}{3\eta_n} \left( \dfrac{\mu_n \left( \lambda_2^2 \xi_{n2}^{-2} - \lambda_1^{-2} \lambda_2^{-2} \xi_{n1}^2 \xi_{n2}^2 \right)}{1 - \left( \lambda_1^2 \xi_{n1}^2 + \lambda_2^2 \xi_{n2}^2 + \lambda_1^{-2} \lambda_2^{-2} \xi_{n1}^2 \xi_{n2}^2 - 3 \right) / J} - \dfrac{\mu_n \left( \lambda_1^2 \xi_{n1}^{-2} - \lambda_1^{-2} \lambda_2^{-2} \xi_{n1}^2 \xi_{n2}^2 \right) / 2}{1 - \left( \lambda_1^2 \xi_{n1}^2 + \lambda_2^2 \xi_{n2}^2 + \lambda_1^{-2} \lambda_2^{-2} \xi_{n1}^2 \xi_{n2}^2 - 3 \right) / J} \right) \end{cases} \tag{20}$$

Based on the theory of nonequilibrium thermodynamics (Suo, 2010), giving an incompressible assumption to the film, the virtual increment component of the free energy density $\delta W_s$ which is contributed by the mechanical force can be expressed as:

$$\delta W_s = \frac{\sigma_1}{\lambda_1} \delta\lambda_1 + \frac{\sigma_2}{\lambda_2} \delta\lambda_2 \tag{21}$$

By adapting the Gent model (Gu et al., 2017b) to the spring of the model for the hyper-elasticity, $W_s$ can be expressed as:

$$W_s = -\frac{\mu_1 J}{2} \ln\left( 1 - \frac{\lambda_1^2 + \lambda_2^2 + \lambda_1^{-2} \lambda_2^{-2} - 3}{J} \right) - \sum_{n=2}^{7} \frac{\mu_n J}{2} \ln\left( 1 - \frac{\lambda_1^2 \xi_{n1}^{-2} + \lambda_2^2 \xi_{n2}^{-2} + \lambda_1^{-2} \lambda_2^{-2} \xi_{n1}^2 \xi_{n2}^2 - 3}{J} \right) \tag{22}$$

By combining Eq (21) and (22), the correlation between the stress and strain can be expressed as:

$$\begin{cases} \sigma_1 = \dfrac{\mu_1\left(\lambda_1^2 - \lambda_1^{-2}\lambda_2^{-2}\right)}{1-\left(\lambda_1^2+\lambda_2^2+\lambda_1^{-2}\lambda_2^{-2}-3\right)\big/J} + \displaystyle\sum_{n=2}^{7} \dfrac{\mu_n\left(\lambda_1^2\xi_{n1}^{-2} - \lambda_1^{-2}\lambda_2^{-2}\xi_{n1}^2\xi_{n2}^2\right)}{1-\left(\lambda_1^2\xi_{n1}^{-2}+\lambda_2^2\xi_{n2}^{-2}+\lambda_1^{-2}\lambda_2^{-2}\xi_{n1}^2\xi_{n2}^2-3\right)\big/J} \\[4mm] \sigma_2 = \dfrac{\mu_1\left(\lambda_2^2 - \lambda_1^{-2}\lambda_2^{-2}\right)}{1-\left(\lambda_1^2+\lambda_2^2+\lambda_1^{-2}\lambda_2^{-2}-3\right)\big/J} + \displaystyle\sum_{n=2}^{7} \dfrac{\mu_n\left(\lambda_2^2\xi_{n2}^{-2} - \lambda_1^{-2}\lambda_2^{-2}\xi_{n1}^2\xi_{n2}^2\right)}{1-\left(\lambda_1^2\xi_{n1}^{-2}+\lambda_2^2\xi_{n2}^{-2}+\lambda_1^{-2}\lambda_2^{-2}\xi_{n1}^2\xi_{n2}^2-3\right)\big/J} \end{cases}$$
(23)

*4.  Summary: the prediction of the inverse behaviour and end effector output force*

Reviewing the algorithm of the flow for dynamic analysis shown in Fig. 4  (B): the transformation from task space to the configuration space is done by the inverse kinematics $K_{kDEA}^{inv}$, where $K_{kDEA}^{inv}$ can be obtained through the combination of Eq (4), (9-11), (17); the transformation from configuration space to the material space is achieved by the combination of $K_{kl}^{DEA}$, the coupled dynamic behaviour between $\xi_{n1}$ and $\xi_{n2}$, and the correlation between the film stress and strain, which respectively are Eq. (4), Eq. (20), and Eq. (23); the transformation from the configuration space to the actuator space is done by the derivation of the virtual work framework, which can be obtained from the combination of the Eqs. (1)-(3). The spring-back force from the PKM forearms is treated as a part of the external payload, which the virtual work can be obtained by the combination of the Eqs. (9), (12-15), (17 ).

| **Algorithm:** Inverse dynamics for a single chain of robot |
|---|
| **Input:** End effector time-dependent track & payload, and geometrical & material parameters |
| **Output:** Time-dependent voltage signal |
| **1**  *start* |
| **2**  *Input material geometric parameter $L_1$, $L_2$ & $L_3$, forearm spring back torque $T(\theta_{kFA})$ (k = 1,2,3) and PKM frame geometric parameters* |
| **3**  *Input material parameters of the spring J, $\mu_n$, dielectric permittivity $\varepsilon$, and the dashpot relaxation time $T_{n-1}$* |
| **4**  *Input end effector track $\mathbf{u}(t)$, and payload $\mathbf{F}(t)$* |
| **5**  *Solve the inverse kinematics through Eqs (4), (9-11), (17) for material stain $\boldsymbol{\lambda}$* |
| **6**  *Initialize the variables $\boldsymbol{\lambda}$ and $\boldsymbol{\xi}_n$* |
| **7**  *Run the inverse dynamic function with the specified time interval t* |
| **8**   *Get the end effector displacement $\mathbf{u}$ at time instant $t_l$* |
| **9**   *Solve the inverse kinematics through Eqs (4), (9-11), (17) for material stain $\boldsymbol{\lambda}_l$ and parallelogram forearm angles $\theta_{kFA}$ at time instant $t_l$* |
| **10**   *Get the forearm spring-back force T at time instant $t_l$* |
| **11**   *Solve $\xi_{n1}$ and $\xi_{n2}$ with given $\boldsymbol{\lambda}_l$ by integrating Eq (20)* |
| **12**   *Solve $\phi(t)$ through Eq (24)* |
| **13**  *end* |
| **14**  *end* |

**Fig. 6** Inverse dynamics for a single chain in the robot. The input of the model is the end effector payload **F** and the time-dependent track **u**, and the output would be the applied actuation voltage $\phi_k$. For a full analysis of the robot, the algorithm needs to run three times for the three kinematic chains.

| **Algorithm:** Output Force prediction for robot |
|---|
| **Input:** End effector time-dependent track, applied time-dependent voltage signal, and geometrical & material parameters |
| **Output:** End effector time-dependent output force |
| **1**  *start* |
| **2**  *Input material geometric parameter $L_1$, $L_2$ & $L_3$, forearm spring back torque $T(\theta_{kFA})$ (k = 1,2,3) and PKM frame geometric parameters* |
| **3**  *Input material parameters of the spring J, $\mu_n$, dielectric permittivity $\varepsilon$, and the dashpot relaxation time $T_{n-1}$* |
| **4**  *Input end effector track $\mathbf{u}(t)$, and voltage tensor $\boldsymbol{\phi}(t)$* |
| **5**  *Solve the inverse kinematics through Eqs (4), (9-11), (17) for material stain $\boldsymbol{\lambda}$* |
| **6**  *Initialize the variables $\boldsymbol{\lambda}$ and $\boldsymbol{\xi}_n$* |
| **7**  *Run the force prediction function with the specified time interval t* |
| **8**   *Get the end effector displacement $\mathbf{u}$ at time instant $t_l$* |
| **9**   *Solve the inverse kinematics through Eqs (4), (9-11), (17) for material stain $\boldsymbol{\lambda}_l$ and parallelogram forearm angles $\theta_{kFA}$ at time instant $t_l$* |
| **10**   *Solve $\xi_{n1}$ and $\xi_{n2}$ with given $\boldsymbol{\lambda}_l$ by integrating Eq (20)* |
| **11**   *Get the forearm spring-back force $\mathbf{F}_{kORTH}$ at time instant $t_l$ through Eq (14) and (17)* |
| **12**   *Get $F_{lORTH}$ through Eq (25)* |
| **13**   *Get the end effector force $\mathbf{F}_u$ by solving Eq (5), (16)* |
| **14**  *end* |
| **15**  *end* |

**Fig. 7** End effector output force prediction for a single chain in the robot. The inputs of the model are the end effector time-dependent track **u** and voltage $\phi_k$, and the output is the forearm tip force.

Based on the aforementioned summary, $\phi_k$ the actuation voltage in kinematic chain $k$ can be expressed as:

$$
\begin{cases}
\phi = \sqrt{\dfrac{G+H+R}{V}} \\[4pt]
G = -\Big(mg\mathbf{x}^j \cdot \mathbf{P}_k^j \mathbf{P}_k^j \mathbf{O}^T + \big(\mathbf{P}_k^j \cdot \mathbf{F}_{k\mathrm{ORTH}}^j\big)\mathbf{P}_k^j \mathbf{O}^T + \big(\mathbf{P}_k^{j\,\cdot\,j+1}\mathbf{M}_j \mathbf{F}_{(k+1)\mathrm{ORTH}}^{j+1}\big)\mathbf{P}_k^j \mathbf{O}^T\Big) \\[4pt]
\quad -\Big(\big(\mathbf{P}_k^{j\,\cdot\,j+2}\mathbf{M}_j \mathbf{F}_{(k+2)\mathrm{ORTH}}^{j+2}\big)\mathbf{P}_k^j \mathbf{O}^T + m_{\mathrm{BC1}}g\mathbf{x}^j \cdot \big(\nabla K_{Qk}^{\mathrm{inv}} \odot \big(\mathbf{P}_k^j \mathbf{P}_k^j \mathbf{O}^T\big)\big)\Big) \\[4pt]
H = -\Big(\big(m\mathbf{P}_k^j \cdot {}^0\mathbf{M}_j \ddot{\mathbf{u}}^0\big)\mathbf{P}_k^j \mathbf{O}^T + \big(m_{\mathrm{BC1}}\nabla^2 K_{Qk}^{\mathrm{inv}} \cdot \mathbf{P}_k^j \mathbf{P}_k^j \cdot {}^0\mathbf{M}_j \ddot{\mathbf{u}}^0\big)\big(\nabla K_{Qk}^{\mathrm{inv}} \cdot \big(\mathbf{P}_k^j \mathbf{P}_k^j \mathbf{O}^T\big)\big)\Big) \\[4pt]
R = L_1 L_2 L_3 \dfrac{\partial K_{k1}^{\mathrm{DEA}}}{\partial \theta_k^{\mathrm{DEA}}} \nabla K_{k\mathrm{DEA}}^{\mathrm{inv}} \cdot \big(\mathbf{P}_k^j \mathbf{P}_k^j \mathbf{O}^T\big)\left(\dfrac{\mu_1\big(\lambda_1^2 - \lambda_1^{-2}\lambda_2^{-2}\big)}{1-\big(\lambda_1^2+\lambda_2^2+\lambda_1^{-2}\lambda_2^{-2}-3\big)\big/J_1} + \sum_{i=2}^{n} \dfrac{\mu_i\big(\lambda_1^2\xi_{i1}^{-2} - \lambda_1^{-2}\lambda_2^{-2}\xi_{i1}^2\xi_{i2}^2\big)}{1-\big(\lambda_1^2\xi_{i1}^{-2}+\lambda_2^2\xi_{i2}^{-2}+\lambda_1^{-2}\lambda_2^{-2}\xi_{i1}^2\xi_{i2}^2-3\big)\big/J_i}\right) \\[4pt]
\quad L_1 L_2 L_3 \dfrac{\partial K_{k2}^{\mathrm{DEA}}}{\partial \theta_k^{\mathrm{DEA}}} \nabla K_{k\mathrm{DEA}}^{\mathrm{inv}} \cdot \big(\mathbf{P}_k^j \mathbf{P}_k^j \mathbf{O}^T\big)\left(\dfrac{\mu_1\big(\lambda_2^2 - \lambda_1^{-2}\lambda_2^{-2}\big)}{1-\big(\lambda_1^2+\lambda_2^2+\lambda_1^{-2}\lambda_2^{-2}-3\big)\big/J_1} + \sum_{i=2}^{n} \dfrac{\mu_i\big(\lambda_2^2\xi_{i2}^{-2} - \lambda_1^{-2}\lambda_2^{-2}\xi_{i1}^2\xi_{i2}^2\big)}{1-\big(\lambda_1^2\xi_{i1}^{-2}+\lambda_2^2\xi_{i2}^{-2}+\lambda_1^{-2}\lambda_2^{-2}\xi_{i1}^2\xi_{i2}^2-3\big)\big/J_i}\right) \\[4pt]
V = \varepsilon\lambda_1\lambda_2 L_1 L_2 L_3^{-1}\left(\lambda_2 \dfrac{\partial K_{k1}^{\mathrm{DEA}}}{\partial \theta_k^{\mathrm{DEA}}} \nabla K_{k\mathrm{DEA}}^{\mathrm{inv}} \cdot \big(\mathbf{P}_k^j \mathbf{P}_k^j \mathbf{O}^T\big) + \lambda_1 \dfrac{\partial K_{k2}^{\mathrm{DEA}}}{\partial \theta_k^{\mathrm{DEA}}} \nabla K_{k\mathrm{DEA}}^{\mathrm{inv}} \cdot \big(\mathbf{P}_k^j \mathbf{P}_k^j \mathbf{O}^T\big)\right)
\end{cases}
\tag{24}
$$

where $\nabla = \left(\frac{\partial}{\partial x}, \frac{\partial}{\partial y}, \frac{\partial}{\partial z}\right)$ is the vector differential operator, $\mathbf{O} = [1, 1, 1]$ is the ones array; $G$, $H$, $R$, and $V$ are the algebraic intermediates where $G$ is the content derived from the external inertia; $H$ is the content derived from the external payload; $R$ is the content derived from the film stretching; $V$ is the content derived from the combination of the applied voltage, electrical field and the electrical participated term in the film constitutive equilibrium. Hence, the algorithm of the inverse dynamics for a single chain in the robot can be summarized in Fig. 6 .

For the end effector force prediction, as shown in Fig. 4 (B): the given end effector track $\mathbf{u}(t)$ from the task space is transformed to the configuration space as $\theta_{k\mathrm{DEA}}(t)$ by the inverse kinematics $K_{k\mathrm{DEA}}^{\mathrm{inv}}$, and then further to the film strains $\lambda_k(t)$ through Eq. (4). Inputting the strains to Eq. (20), the coupled dynamic behaviour between $\xi_{n1}$ and $\xi_{n2}$ could hence be obtained. Then, with the given time-dependent voltage $\phi_k(t)$, the film stress tensor $\sigma_k(t)$ can be obtained through Eq. (23). This allows $F_{k\mathrm{PARA}}$ to be expressed as below:

$$
\begin{cases}
F_{k\mathrm{PARA}} = \dfrac{G+H+R}{V} \\[4pt]
G = -\Big(mg\mathbf{x}^j \cdot \mathbf{P}_k^j \mathbf{P}_k^j \mathbf{O}^T + m_{\mathrm{BC1}}g\mathbf{x}^j \cdot \big(\nabla K_{Qk}^{\mathrm{inv}} \odot \big(\mathbf{P}_k^j \mathbf{P}_k^j \mathbf{O}^T\big)\big)\Big) \\[4pt]
\quad -\Big(\big(\mathbf{P}_k^j \cdot \mathbf{F}_{k\mathrm{ORTH}}^j\big)\mathbf{P}_k^j \mathbf{O}^T + \big(\mathbf{P}_k^{j\,\cdot\,j+1}\mathbf{M}_j \mathbf{F}_{(k+1)\mathrm{ORTH}}^{j+1}\big)\mathbf{P}_k^j \mathbf{O}^T + \big(\mathbf{P}_k^{j\,\cdot\,j+2}\mathbf{M}_j \mathbf{F}_{(k+2)\mathrm{ORTH}}^{j+2}\big)\mathbf{P}_k^j \mathbf{O}^T\Big) \\[4pt]
\quad -\Big(\big(m\mathbf{P}_k^j \cdot {}^0\mathbf{M}_j \ddot{\mathbf{u}}^0\big)\mathbf{P}_k^j \mathbf{O}^T + \big(m_{\mathrm{BC1}}\nabla^2 K_{Qk}^{\mathrm{inv}} \cdot \mathbf{P}_k^j \mathbf{P}_k^j \cdot {}^0\mathbf{M}_j \ddot{\mathbf{u}}^0\big)\big(\nabla K_{Qk}^{\mathrm{inv}} \cdot \big(\mathbf{P}_k^j \mathbf{P}_k^j \mathbf{O}^T\big)\big)\Big) \\[4pt]
H = -\varepsilon\phi^2 \lambda_1\lambda_2 L_1 L_2 L_3^{-1}\left(\lambda_2 \dfrac{\partial K_{k1}^{\mathrm{DEA}}}{\partial \theta_k^{\mathrm{DEA}}} \nabla K_{k\mathrm{DEA}}^{\mathrm{inv}} \cdot \big(\mathbf{P}_k^j \mathbf{P}_k^j \mathbf{O}^T\big) + \lambda_1 \dfrac{\partial K_{k2}^{\mathrm{DEA}}}{\partial \theta_k^{\mathrm{DEA}}} \nabla K_{k\mathrm{DEA}}^{\mathrm{inv}} \cdot \big(\mathbf{P}_k^j \mathbf{P}_k^j \mathbf{O}^T\big)\right) \\[4pt]
R = L_1 L_2 L_3 \dfrac{\partial K_{k1}^{\mathrm{DEA}}}{\partial \theta_k^{\mathrm{DEA}}} \nabla K_{k\mathrm{DEA}}^{\mathrm{inv}} \cdot \big(\mathbf{P}_k^j \mathbf{P}_k^j \mathbf{O}^T\big)\left(\dfrac{\mu_1\big(\lambda_1^2 - \lambda_1^{-2}\lambda_2^{-2}\big)}{1-\big(\lambda_1^2+\lambda_2^2+\lambda_1^{-2}\lambda_2^{-2}-3\big)\big/J_1} + \sum_{i=2}^{n} \dfrac{\mu_i\big(\lambda_1^2\xi_{i1}^{-2} - \lambda_1^{-2}\lambda_2^{-2}\xi_{i1}^2\xi_{i2}^2\big)}{1-\big(\lambda_1^2\xi_{i1}^{-2}+\lambda_2^2\xi_{i2}^{-2}+\lambda_1^{-2}\lambda_2^{-2}\xi_{i1}^2\xi_{i2}^2-3\big)\big/J_i}\right) \\[4pt]
\quad + L_1 L_2 L_3 \dfrac{\partial K_{k2}^{\mathrm{DEA}}}{\partial \theta_k^{\mathrm{DEA}}} \nabla K_{k\mathrm{DEA}}^{\mathrm{inv}} \cdot \big(\mathbf{P}_k^j \mathbf{P}_k^j \mathbf{O}^T\big)\left(\dfrac{\mu_1\big(\lambda_2^2 - \lambda_1^{-2}\lambda_2^{-2}\big)}{1-\big(\lambda_1^2+\lambda_2^2+\lambda_1^{-2}\lambda_2^{-2}-3\big)\big/J_1} + \sum_{i=2}^{n} \dfrac{\mu_i\big(\lambda_2^2\xi_{i2}^{-2} - \lambda_1^{-2}\lambda_2^{-2}\xi_{i1}^2\xi_{i2}^2\big)}{1-\big(\lambda_1^2\xi_{i1}^{-2}+\lambda_2^2\xi_{i2}^{-2}+\lambda_1^{-2}\lambda_2^{-2}\xi_{i1}^2\xi_{i2}^2-3\big)\big/J_i}\right) \\[4pt]
V = \mathbf{n}_k^j \cdot \mathbf{P}_k^j \mathbf{P}_k^j \mathbf{O}^T
\end{cases}
\tag{25}
$$

where $G$, $H$, $R$, and $V$ are the algebraic intermediates where $G$ is the content from the electrical field; $H$ is the content from the film stretching; R is the content from the forearm spring-back force in chain $k$; $V$ is the displacement projection from the world to the direction that paralleled to the forearm. Hence, the algorithm of the force prediction for a single chain in the robot can be summarized in Fig. 7.

## IV. EXPERIMENTAL RESULTS

This section provides the ageing behaviour of the double-layered carbon paint electrode (DCPE) and the validation of the proposed inverse model.

The test of the electrode ageing behaviour focuses on the dynamic response of a constructed DEA sample. As a comparison, a dynamic response of a DEA sample made with carbon grease electrode was also recorded. To validate the proposed inverse model, we constructed a prototype, and tested the agreement of the proposed circular paths with the real path under a motion capture under a calibration of the film constitutive parameters and forearm spring-back torque. Additionally, the output force prediction function is validated by placing a dead weight on the prototype actuated by the prefined voltage.

### 1. Time-dependent stability test of the electrode

To compare the behaviour between the proposed paint and the carbon grease, uniaxially tensioned DEA samples have been built as shown in Fig. 8. Two ends of the dielectric film (Elastosil 2030, Silex Technology Inc.) are fixed to strips with a neutral dimension of 45mm×60mm×0.1mm to form a DEA sample. To avoid the short circuit, a 3mm gap has been left on each side of the film when painting the electrode. The two samples are hung on a test rig and respectively take a 50g payload at the moving end as shown in Fig. 8(B). The signals are all generated by a microcontroller (Mega 2560, Arduino) and respectively amplified by a bipolar voltage controller (HVA0560, HVM Technology Inc.) for each output channel as shown in Fig. 8(C, D). The time-dependent movements of each film are tracked by the markers that are fixed on the end of the samples through a motion capture system (X16, Vicon Motion Systems Ltd). Carbon grease electrode (CGE) and double-layered carbon paint electrode (DCPE) are respectively applied to the DEA samples. Where the weight ratio between the carbon black, the photosensitive resin, and IPA is 81:129:764 and the electrodes are cured in the curing box under 40° C for 600 minutes.

A comparison between the behavior of the two electrodes is shown in Fig. 9 . The samples are driven by a 1/3Hz signal with

2.5kV amplitude and a 2.5kV biasing (i.e. wave crest 5kV and wave through 0V). The two samples were tested right after the fabrication as the behaviour of day 1, then kept under a temperature between 20~23°C and relative humidity (RH) between 38~42% and tested on days 3, 7, and 14 for the comparison of their longer-term behaviour. Focusing on the behaviour of the first eight loops of the DEA samples, comparing the first day data, the CGE sample observed a 0.23mm decrement (17.5%) of its average stroke at day 3, a 0.49mm decrement (38.4%) at day 7, and a 0.82mm decrement (64.3%) at day 14. For the DCPE sample, the variation of the average stroke is under 0.086mm (6%) which is partly caused by measuring errors.

It should be noted that the test focuses on the variation of the electrode caused by the electrode ageing. The elongation between the CGE and DCPE samples is incomparable as the film of the DCPE sample was scanned by the laser, hence the effective thickness of the film on the DCPE sample is slightly lower than the film on the CGE sample. Further, the behaviour of the CGE sample can have a huge difference according to the applied thickness of the electrode. A thicker electrode offers a slower aging, but the tradeoff is easier contamination to other unapplied areas, and a worse control when constructing multi-layered actuators in the future. In this test, we applied a relatively thin electrode thickness which gives a faster testing result but still remains representative of an overall trend of carbon grease ageing.

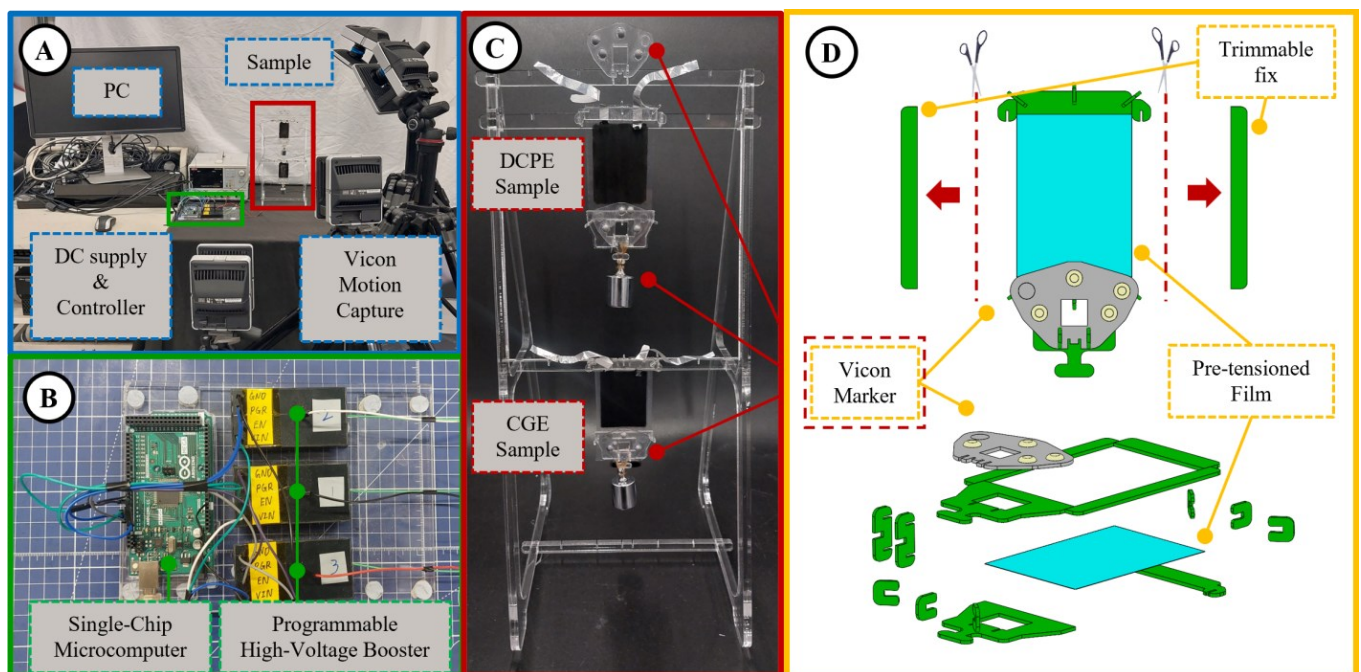

**Fig. 8** Test of the electrode dynamic behaviour along the ageing. (A) The setup of the test. The proposed double-layered carbon black paint electrode (DCPE) and the general carbon grease electrode (CGE) are respectively applied to silicone film to form DEA samples to test the behaviour. The two samples are hung on a frame and the free ends are loaded with 50g weight. The displacement of the DEAs is obtained by a motion-capturing system (Vicon). (B) The controller set. A single-chip microcomputer generates pre-programmed pulse-width modulation (PWM) signals and feeds into three programmable high-voltage amplifiers. This provides a maximum of 3 channel control to the mounted prototype. (C) The prototype for the test. To track the movement, each free end of the DEA is mounted with a marker, and an additional marker is mounted on the frame to set a global origin for the tracking. (D) The design of the testing sample. The films are fixed to acrylic strips to form a testing sample and the two horizontal ends of the sample are free. Same with the construction of the DEA for PKM, the frames are temporarily locked by trimmable fixes during the construction to guarantee a consistent length of the sample, then are unlocked during the real test.

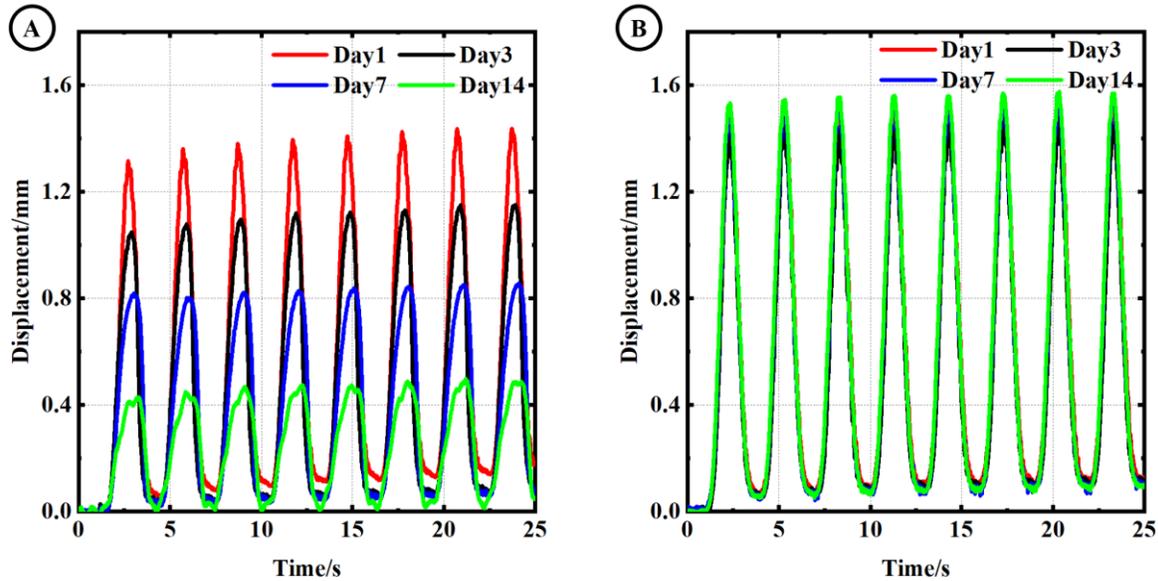

**Fig. 9** Dynamic results of the DEA samples driven by 1/3Hz sin signal with an amplitude of 2.5kV and a 2.5kV biasing. The samples were tested right after their construction (Day 1) and then were tested in sequence after a standing of 3 days, 7 days, and 14 days. (A) The result of the DEA sample was applied with grease electrodes (CGE). (B) The result of the DEA sample was applied with double-layered carbon black paint electrodes (DCPE).

### 2. Setup, and calibration of the DEA-based PKMl

Following the process mentioned in section II, we constructed the DEA-based PKM with the key parameters shown in Tab 1 with the same dielectric film (Elastosil 2030, Silex Technology Inc.) to the DCPE sample, 5mm acrylic sheet for the bottom base (Mechanical ground), 0.5mm PETG for the PKM forearm, and 1mm PETG for the rest of the PKM components. The total weight of the PKM is 54.5g excluding the mechanical ground.

**Tab 1** Key parameters for the construction of the PKM

| | |
|---|---|
| $m$ | 6g |
| $m_{bc}$ | 0.96g |
| $a$ | 55mm |
| $b$ | 10mm |
| $c$ | 11.5mm |
| $d$ | 62mm |
| $q$ | 50mm |
| $n_0$ | 50mm |
| $l_{DEA}$ | 50mm |
| $l_{ch}$ | 30mm |
| $l_{cv}$ | 82.39mm |
| $l_{end}$ | 25mm |
| $\omega$ | 90° |
| $\alpha$ | 3° |
| $\theta_{kDEA}$ | 111.8° |
| $L_1$ | 62.9mm |
| $L_2$ | 41.5mm |
| $L_3$ | 0.085mm |
| $\theta_{pre}$ | 141.23° |
| $\lambda_{pre}$ | 1.5 |

Similar to the test setup mentioned in section IV.1, we validated the effectiveness of the proposed modelling through the motion capture (Fig. 10). The films of the PKM actuators were first calibrated by simultaneously applying a sin signal with 2.75kV amplitude and a 2.75kV biasing. The motion of the moving end effector was recorded by motion capture and then imported into the inverse model to predict the actuated voltage. By approaching the predicted voltage to the real voltage, the constitutive model given in section III.3 can be determined. Further, by applying the same signal to a single actuator, the spring-back torque from each of the forearms was calibrated and the result is shown in Fig. 10 (D).

A considerable difference was observed during the calibration of the driven voltage below 1.5kV. This is due to the resistance of the conductive wires and electrodes as well as the relatively high film thickness during a small elongation. However, thanks to the kinematic structure of the parallel mechanism, the actuators require a higher stroke to move the end effector when the end effector is in the upper area of the workspace. This, as a result, compensates for a part of the inaccuracy during the low-driven voltage area.

### 3. Validation of the DEA-based PKM inverse model

As shown in Fig. 11, circular paths were generated to validate the proposed inverse model. The end effector was first descended to the specified height and then placed to the start point (right side) of the pattern. The end effector was then looped 5 cycles according to the specified pattern and finally ended the test after moving back to the neutral position. The motion of the end effector was captured under a sampling rate of 100 Hz. For the pattern of a 1/3 Hz circle with no payload, the maximum error is 0.56mm (5.6%), -0.52mm (-5.2%) and 2.03mm (6.0%) respectively on the x, y and z axis, and the root mean square error (RMSE) is 0.18 (1.8%), 0.21 (2.1%), and 1.03mm (3.0%). It should be noted that Fig. 11 (C), (F) and (I) show the value based on the prefined world coordinated shown in Fig. 4 (A). And the relative error for x and y axis are based on the diameter of the circular pattern, and z axis is based on the maximum vertical stroke under the correlated applied payload (34mm for zero payload and 49.3mm for 2g payload). For the pattern of 1/3 Hz circle with 2g payload (9.6mm diameter), the maximum error is 0.39mm (4.1%), 0.67mm (7.0%) and 1.44mm (3.0%) respectively on the x, y and z axis, and the RMSE are 0.13 (1.3%), 0.24 (2.5%), and 0.81mm (1.6%). And for the pattern of 1/8 Hz circle with no payload, the maximum error is -0.67mm (-6.7%), -0.50mm (-5.0%) and -1.93mm (-5.6%) respectively on the x, y and z axis, and the RMSE are 0.23 (1.7%), 0.18 (2.2%), and 0.81mm (2.3%).

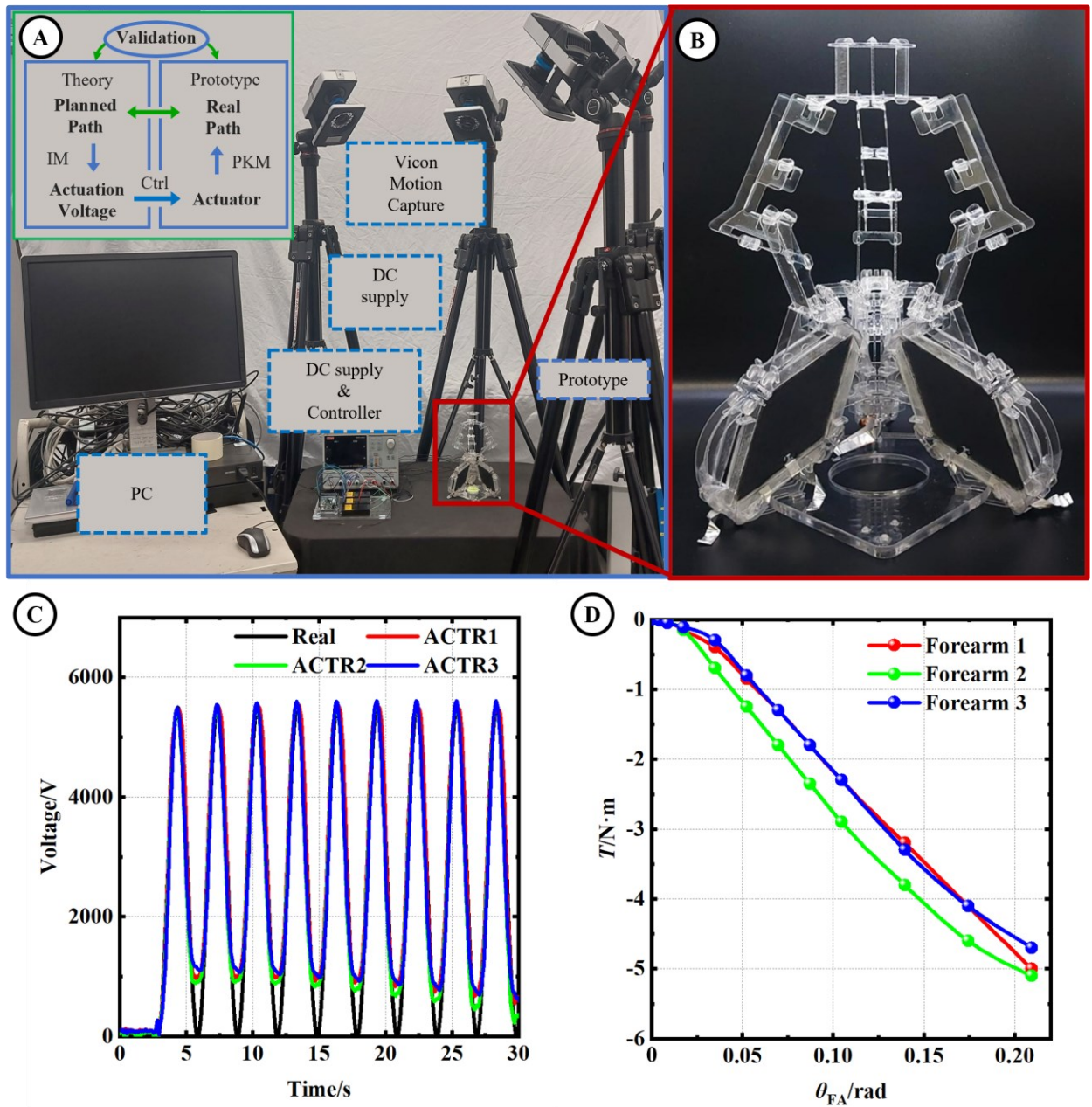

**Fig. 10** Setup and calibration of the DEA-based PKM. (A) The setup for the calibration and validation. The setup is the same as with Fig. 8. For the validation of the model, the planned path is input into the inverse model (IM) and the actuation voltage is obtained as an output. By feeding the actuation signal into the controller and finally enabling the parallel kinematic mechanism (PKM), the real motion is obtained through a motion capture system (Vicon). The validation of the inverse model is done by the comparison between the planned path and the real path. (B) The prototype of the designed PKM. The prototype consists of two Vicon markers respectively on the end effector and the frame of the robot. (C) The calibration result of the film. A sine wave with 1/3Hz, 2.75kV amplitude and a 2.75kV biasing was supplied to the three actuators simultaneously. The figure shows the comparison of the time-dependent voltage between the model-predicted result and the prefined signal. (D) The calibrated forearm torsional joint behaviour. Keeping with the same input signal in (C), the calibration of the torsional joint was done by actuating one of the actuators and keeping the rest of the two unactuated. Several angles were manually selected and prefined with a spring-back torque. Treating the interpolation of the prefined scatters as the behaviour of the torsional joints, the final behaviour was determined by fitting the predicted driven voltage with the calibrated voltage shown in (C).

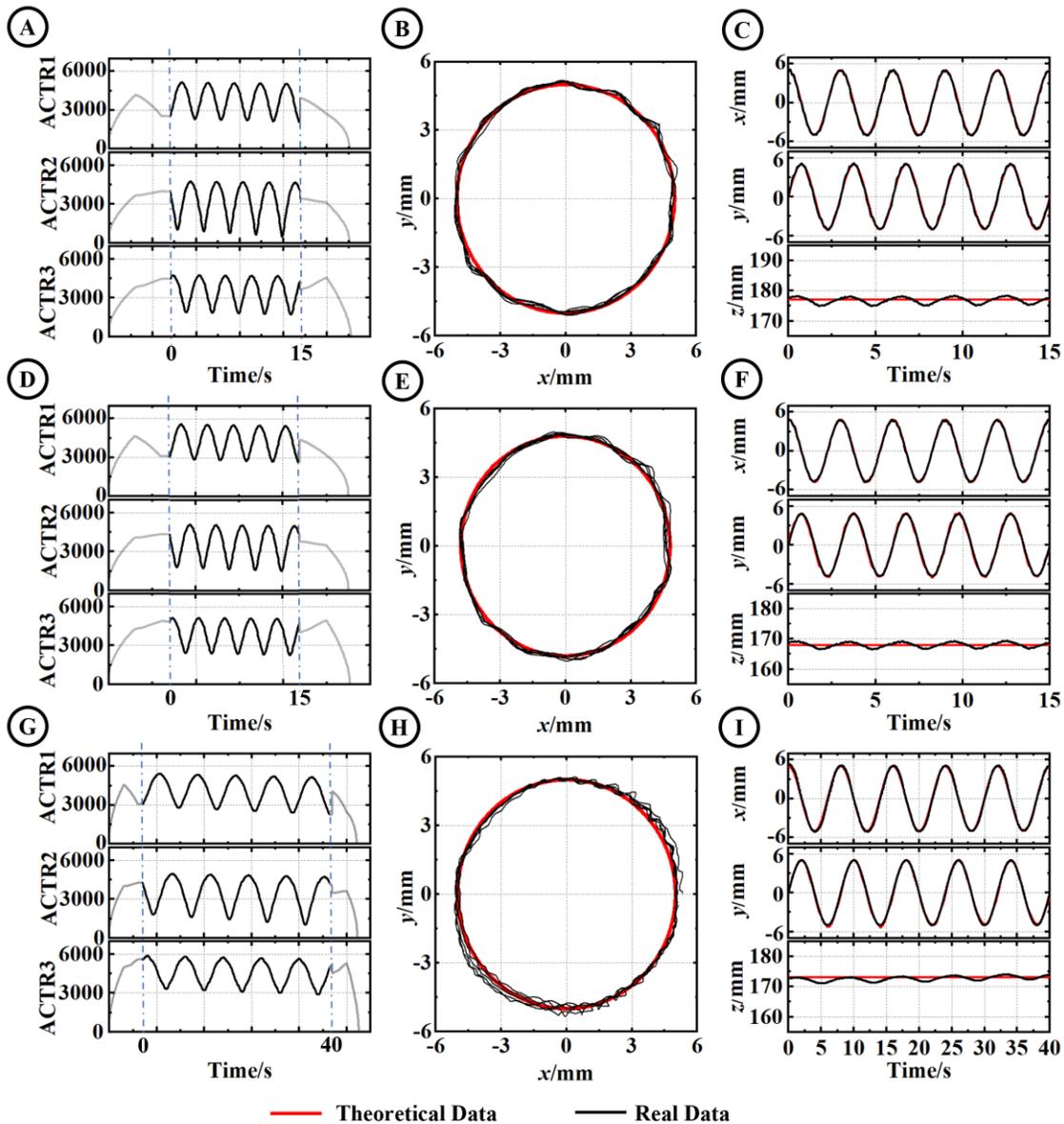

**Fig. 11** Result of the inverse model validation. (A), (D) and (G) are the input voltage of the three actuators for paths correspondingly with 3s per cycle with zero payload, 3s per cycle with 2g payload, and 8s per circle with zero payload. The transparent parts are to set the PKM end effector to the start point of the prefined path and the end of the prefined path. (B), (E) and (H) are the top view of the prefined path corresponding to the input voltage in (A), (D) and (G). (C), (F) and (I) are the dynamic response of the circular path corresponding to the input voltage in (A) in the x, y and z axis.

## 4. Output force prediction of the DEA-based PKM end effector

This subsection records the test of the PKM output force prediction. With the same setup mentioned in Fig. 8 and Fig. 10, we placed a 5g weight on the end effector and further moved the end effector up and down with a prefined driving voltage Fig. 12 (A). The test starts with zero payload and then manually places a 5g (i.e. 0.05N) weight on the end effector. Feeding the tracked end effector position to the model, the predicted payload in z axis and the value of the resultant force in xy plane are respectively shown in Fig. 12 (B) and (C).

According to Fig. 12(B), for the holding period during the actuation, the prediction in z axis shows an RMSE of 0.006N (12.4%) with a maximum error of 0.016N. A larger error (RSME 0.023N) was observed after the driving voltage decreased to zero and this is potentially caused by the approaching the PKM workspace boundary and the resistance from the PKM joint. The period when the end effector moves show an RMSE of 0.07N with a maximum error of -0.33N, which appears at the moment of 46.5s, which is the moment that the end effector starts to recover to its unactuated state and induces a bouncing movement. Fig. 12(C) shows the value of the resultant force in the xy plane. The RMSE for the holding period during the actuation is 0.02N and the RMSE for the end effector moving period is 0.03N, which is caused by the error from the manufacturing of the kinematic structure in DEA-based PKM.

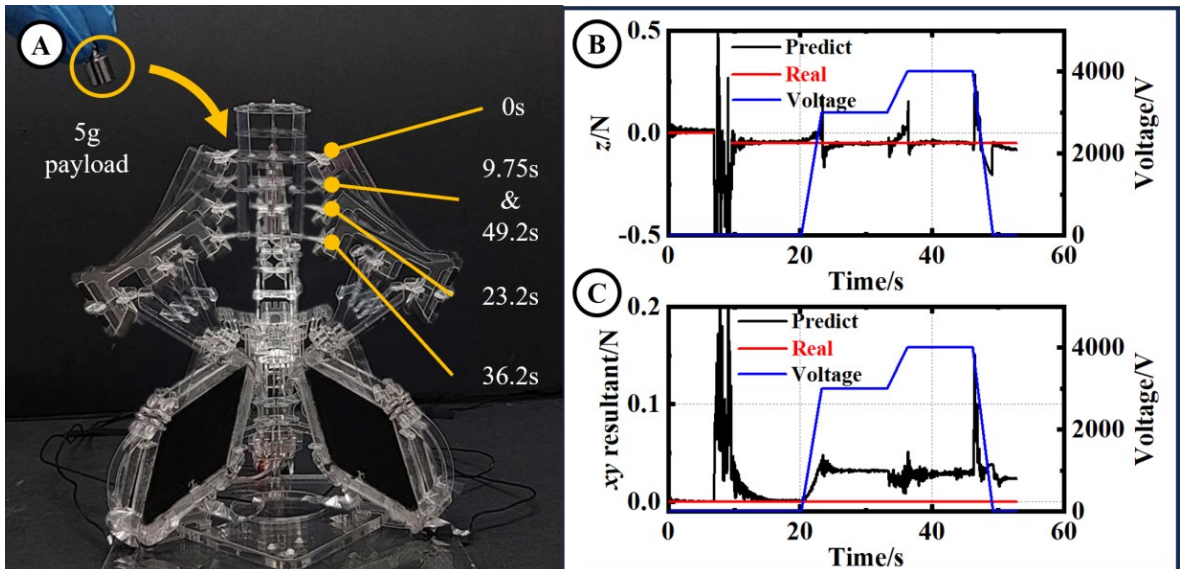

**Fig. 12** Prediction of the DEA-based PKM end effector output force: 0-6.95s is the period with zero payload and no driving voltage; 6.95-20.2s is the period we manually placed a 5g weight to the PKM end effector and wait until it stabled (at ~12s); 20.2-23.2s is the period that we started to add a linearly increasing driving voltage from 0V to 3000V and then hold 10s for stability test (period of 23.2-33.2s); the driven voltage in the period of 33.2-36.2s was further increased linearly to 4000V and then again hold 10s for stability test (period of 36.2-46.2s); the period 46.2-49.2s linearly decreases the driving voltage back to 0 and finally ends the test at 52.8s. (A) The snapshot of the tests (B) The predicted payload vs time in z axis (with direction; Black line- predicted force, Red Line – real payload, Blue line - Voltage). (C) The value of the resultant force of predicted payload vs time in xy plane.

## V. DISCUSSION AND CONCLUSION

We present the design, fabrication, modelling, and control of a parallel kinematics robot (also known as PKM) actuated by multiple dielectric elastomer actuators (DEAs). We introduced an innovative double-layered carbon black paint electrode (DCPE) that applies the carbon black to the laser-scanned dielectric film and is protected by a mixture of carbon black and photosensitive resin, significantly enhancing the electrode's stability and adherence compared to the general carbon grease electrode. This method improves the long-term behaviour of the electrodes, decreases the contamination risk caused by the carbon grease, and potentially simplifies the control of the electrode layer thickness. Based on the DEA-actuated Delta PKM, we develop an inverse dynamic model that incorporates an expanded Bergstrom-Boyce model into the constitutive behavior of the dielectric film. This model can accurately predicts the output force and the inverse motion of the robot's end effector, demonstrating a predictive maximum relative RSME of 2.5% for the path tracking and a relative RSME of 12.4% in the holding period for the output force prediction.

The use of lozenge shape DEAs (LSDEAs) in the PKM offers a more uniform distribution of mechanical stress and friendlier distribution for DEA with multiple layers. The proposed paint method for electrode preparation uses a mixture of carbon black and photosensitive resin. This approach significantly improves the stability and repeatability of the electrode's dynamic behavior, enhancing DEA performance by reducing the variation of the electrode stiffness and conductivity caused by dehydration and ageing. The incorporation of SAMs into the DEA design offers a route of retaking the trade-off between the output force and stroke, which from the soft robot's perspective enhanced the PKM's operational range. By employing 3D puzzling strip structures for the construction of the PKM, the assembly process is simplified. This innovative approach decreases the robot's weight while maintaining structural integrity, facilitating easier and more cost-effective manufacturing. This research demonstrates effective collaboration between multiple DEAs to enable precise inverse dynamics and output force predictions through a generic inverse model. This provides a numerical safety evaluation to the extra delicate operation both during normal work and when an unexpected accidental collision happens.

This work contributes to a model of collaboration between multiple electroactive polymers, and the construction of a stabled, stroke-amplified actuator. This engaged the potentially higher integration and precision soft actuators into the use of collaborative manipulations. To broaden the applicability in more sensitive and rapid operational contexts, future work earmarks the sensing of the robot, a resonance-considered rapid actuation model and a higher robot performance of stroke and payload taking.

### ACKNOWLEDGEMENTS

This work was supported by the university of Nottingham and Innovate UK (51689).

### CONFLICT OF INTEREST

The authors declare no conflict of interest.

REFERENCES

Atia MG, Mohammad A, Gameros A, Axinte D and Wright I (2022) Reconfigurable Soft Robots by Building Blocks. *Advanced Science.* 2203217.

Axinte D, Dong X, Palmer D, Rushworth A, Guzman SC, Olarra A, Arizaga I, Gomez-Acedo E, Txoperena K and Pfeiffer K (2018) MiRoR—Miniaturized robotic systems for holistic in-situ repair and maintenance works in restrained and hazardous environments. *IEEE/ASME Transactions on Mechatronics* 23(2): 978-981.

Baltes M, Kunze J, Prechtl J, Seelecke S and Rizzello G (2022) A bi-stable soft robotic bendable module driven by silicone dielectric elastomer actuators: design, characterization, and parameter study. *Smart Materials and Structures* 31(11): 114002.

Bergström JS and Boyce M (1998) Constitutive modeling of the large strain time-dependent behavior of elastomers. *Journal of the Mechanics and Physics of Solids* 46(5): 931-954.

Black CB, Till J and Rucker DC (2017) Parallel continuum robots: Modeling, analysis, and actuation-based force sensing. *IEEE Transactions on Robotics* 34(1): 29-47.

Borenstein J (1995) Control and kinematic design of multi-degree-of freedom mobile robots with compliant linkage. *IEEE transactions on robotics and automation* 11(1): 21-35.

Camacho-Arreguin JI, Wang M, Russo M, Dong X and Axinte D (2022) Novel Reconfigurable Walking Machine Tool Enables Symmetric and Nonsymmetric Walking Configurations. *IEEE/ASME Transactions on Mechatronics* 27(6): 5495-5506.

De Payrebrune KM and O'Reilly OM (2017) On the development of rod-based models for pneumatically actuated soft robot arms: A five-parameter constitutive relation. *International Journal of Solids and Structures* 120: 226-235.

Dong X, Wang M, Mohammad A, Ba W, Russo M, Norton A, Kell J and Axinte D (2022) Continuum robots collaborate for safe manipulation of high-temperature flame to enable repairs in challenging environments. *IEEE/ASME Transactions on Mechatronics* 27(5): 4217-4220.

Duduta M, Wood RJ and Clarke DR (2016) Multilayer dielectric elastomers for fast, programmable actuation without prestretch. *Advanced Materials* 28(36): 8058-8063.

Fraś J, Czarnowski J, Maciaś M and Główka J (2014) Static modeling of multisection soft continuum manipulator for stiff-flop project. *Recent Advances in Automation, Robotics and Measuring Techniques.* Springer, 365-375.

Garcia M, Moghadam AAA, Tekes A and Emert R (2020) Development of a 3D printed soft parallel robot. *ASME International Mechanical Engineering Congress and Exposition.* American Society of Mechanical Engineers, V07AT07A008.

Grazioso S, Di Gironimo G and Siciliano B (2019) A geometrically exact model for soft continuum robots: The finite element deformation space formulation. *Soft robotics* 6(6): 790-811.

Gu G-Y, Gupta U, Zhu J, Zhu L-M and Zhu X (2017a) Modeling of viscoelastic electromechanical behavior in a soft dielectric elastomer actuator. *IEEE Transactions on Robotics* 33(5): 1263-1271.

Gu G-Y, Zhu J, Zhu L-M and Zhu X (2017b) A survey on dielectric elastomer actuators for soft robots. *Bioinspiration & biomimetics* 12(1): 011003.

Gu G, Zou J, Zhao R, Zhao X and Zhu X (2018) Soft wall-climbing robots. *Science Robotics* 3(25): eaat2874.

Guo Y, Liu L, Liu Y and Leng J (2021) Review of dielectric elastomer actuators and their applications in soft robots. *Advanced Intelligent Systems* 3(10): 2000282.

Hau S, Rizzello G and Seelecke S (2018) A novel dielectric elastomer membrane actuator concept for high-force applications. *Extreme Mechanics Letters* 23: 24-28.

Hodgins M, York A and Seelecke S (2011) Modeling and experimental validation of a bi-stable out-of-plane DEAP actuator system. *Smart Materials and Structures* 20(9): 094012.

Hongzhe Z and Shusheng B (2010) Accuracy characteristics of the generalized cross-spring pivot. *Mechanism and Machine Theory* 45(10): 1434-1448.

Huang X, Zhu X and Gu G (2022) Kinematic modeling and characterization of soft parallel robots. *IEEE Transactions on Robotics* 38(6): 3792-3806.

Iqbal J, Khan ZH and Khalid A (2017) Prospects of robotics in food industry. *Food Science and Technology* 37: 159-165.

Ji X, El Haitami A, Sorba F, Rosset S, Nguyen GT, Plesse C, Vidal F, Shea HR and Cantin S (2018) Stretchable composite monolayer electrodes for low voltage dielectric elastomer actuators. *Sensors and Actuators B: Chemical* 261: 135-143.

Ji X, Liu X, Cacucciolo V, Imboden M, Civet Y, El Haitami A, Cantin S, Perriard Y and Shea H (2019) An autonomous untethered fast soft robotic insect driven by low-voltage dielectric elastomer actuators. *Science Robotics* 4(37): eaaz6451.

Kozuka H, Arata J, Okuda K, Onaga A, Ohno M, Sano A and Fujimoto H (2013) A compliant-parallel mechanism with bio-inspired compliant joints for high precision assembly robot. *Procedia Cirp* 5: 175-178.

Krajcovic M, Bulej V, Kuric I and Sapietova A (2013) Intelligent manufacturing systems in concept of digital factory. *Communications-Scientific Letters of the University of Zilina* 15(2): 77-87.

Laski PA, Takosoglu JE and Blasiak S (2015) Design of a 3-DOF tripod electro-pneumatic parallel manipulator. *Robotics and Autonomous Systems* 72: 59-70.

Li C, Gu X, Xiao X, Lim CM and Ren H (2018) A robotic system with multichannel flexible parallel manipulators for single port access surgery. *IEEE Transactions on Industrial Informatics* 15(3): 1678-1687.

Lindenroth L, Housden RJ, Wang S, Back J, Rhode K and Liu H (2019) Design and integration of a parallel, soft robotic end-effector for extracorporeal ultrasound. *IEEE Transactions on Biomedical Engineering* 67(8): 2215-2229.

Lindenroth L, Soor A, Hutchinson J, Shafi A, Back J, Rhode K and Liu H (2017) Design of a soft, parallel end-effector applied to robot-guided ultrasound interventions. *2017 IEEE/RSJ International Conference on Intelligent Robots and Systems (IROS).* IEEE, 3716-3721.

Ma N, Dong X and Axinte D (2020) Modeling and experimental validation of a compliant underactuated parallel kinematic manipulator. *IEEE/ASME Transactions on Mechatronics* 25(3): 1409-1421.

Ma N, Dong X, Palmer D, Arreguin JC, Liao Z, Wang M and Axinte D (2019) Parametric vibration analysis and validation for a novel portable hexapod machine tool attached to surfaces with unequal stiffness. *Journal of Manufacturing Processes* 47: 192-201.

Majidi C (2014) Soft robotics: a perspective—current trends and prospects for the future. *Soft robotics* 1(1): 5-11.

Marchese AD and Rus D (2016) Design, kinematics, and control of a soft spatial fluidic elastomer manipulator. *The International Journal of Robotics Research* 35(7): 840-869.

Merlet J-P (2006) *Parallel robots.* Springer Science & Business Media.

Moghadam AAA, Kouzani A, Torabi K, Kaynak A and Shahinpoor M (2015) Development of a novel soft parallel robot equipped with polymeric artificial muscles. *Smart Materials and Structures* 24(3): 035017.

Moretti G, Sarina L, Agostini L, Vertechy R, Berselli G and Fontana M (2020) Styrenic-rubber dielectric elastomer actuator with inherent stiffness compensation. *Actuators.* MDPI, 44.

Nakano T, Sugita N, Ueta T, Tamaki Y and Mitsuishi M (2009) A parallel robot to assist vitreoretinal surgery. *International journal of computer assisted radiology and surgery* 4: 517-526.

Oliver-Butler K, Till J and Rucker C (2019) Continuum robot stiffness under external loads and prescribed tendon displacements. *IEEE Transactions on Robotics* 35(2): 403-419.

Park Y-J, Jeong U, Lee J, Kwon S-R, Kim H-Y and Cho K-J (2012) Kinematic condition for maximizing the thrust of a robotic fish using a compliant caudal fin. *IEEE Transactions on Robotics* 28(6): 1216-1227.

Pelrine R, Kornbluh R and Kofod G (2000) High‐strain actuator materials based on dielectric elastomers. *Advanced Materials* 12(16): 1223-1225.

Pelrine R, Sommer-Larsen P, Kornbluh RD, Heydt R, Kofod G, Pei Q and Gravesen P (2001) Applications of dielectric elastomer actuators. *Smart Structures and Materials 2001: Electroactive Polymer Actuators and Devices.* SPIE, 335-349.

Pile J and Simaan N (2014) Modeling, design, and evaluation of a parallel robot for cochlear implant surgery. *IEEE/ASME Transactions on Mechatronics* 19(6): 1746-1755.

Plante J-S and Dubowsky S (2007) On the performance mechanisms of dielectric elastomer actuators. *Sensors and actuators A: Physical* 137(1): 96-109.

Renda F, Boyer F, Dias J and Seneviratne L (2018) Discrete cosserat approach for multisection soft manipulator dynamics. *IEEE Transactions on Robotics* 34(6): 1518-1533.

Renda F, Cianchetti M, Giorelli M, Arienti A and Laschi C (2012) A 3D steady-state model of a tendon-driven continuum soft manipulator inspired by the octopus arm. *Bioinspiration & biomimetics* 7(2): 025006.

Rosset S, Araromi OA, Schlatter S and Shea HR (2016) Fabrication process of silicone-based dielectric elastomer actuators. *JoVE (Journal of Visualized Experiments)*.(108): e53423.

Sarban R, Lassen B and Willatzen M (2011) Dynamic electromechanical modeling of dielectric elastomer actuators with metallic electrodes. *IEEE/ASME Transactions on Mechatronics* 17(5): 960-967.

Shi Y, Askounis E, Plamthottam R, Libby T, Peng Z, Youssef K, Pu J, Pelrine R and Pei Q (2022) A processable, high-performance dielectric elastomer and multilayering process. *Science* 377(6602): 228-232.

Singh I, Amara Y, Melingui A, Mani Pathak P and Merzouki R (2018) Modeling of continuum manipulators using pythagorean hodograph curves. *Soft robotics* 5(4): 425-442.

Sriratanasak N, Axinte D, Dong X, Mohammad A, Russo M and Raimondi L (2022) Tasering Twin Soft Robot: A Multimodal Soft Robot Capable of Passive Flight and Wall Climbing. *Advanced Intelligent Systems*. 2200223.

Suo Z (2010) Theory of dielectric elastomers. *Acta Mechanica Solida Sinica* 23(6): 549-578.

Wang L and Simaan N (2019) Geometric calibration of continuum robots: Joint space and equilibrium shape deviations. *IEEE Transactions on Robotics* 35(2): 387-402.

Wang X, Li S, Chang J-C, Liu J, Axinte D and Dong X (2024) Multimodal locomotion ultra-thin soft robots for exploration of narrow spaces. *Nature Communications* 15(1): 6296.

Wang Y and Xu Q (2021) Design and testing of a soft parallel robot based on pneumatic artificial muscles for wrist rehabilitation. *Scientific Reports* 11(1): 1273.

Webster III RJ and Jones BA (2010) Design and kinematic modeling of constant curvature continuum robots: A review. *The International Journal of Robotics Research* 29(13): 1661-1683.

Wissler M and Mazza E (2007) Electromechanical coupling in dielectric elastomer actuators. *Sensors and actuators A: Physical* 138(2): 384-393.

Xu Z, Zheng S, Wu X, Liu Z, Bao R, Yang W and Yang M (2019) High actuated performance MWCNT/Ecoflex dielectric elastomer actuators based on layer-by-layer structure. *Composites Part A: Applied Science and Manufacturing* 125: 105527.

Youn J-H, Jeong SM, Hwang G, Kim H, Hyeon K, Park J and Kyung K-U (2020) Dielectric elastomer actuator for soft robotics applications and challenges. *Applied Sciences* 10(2): 640.

Zhao H, Bi S and Yu J (2012) A novel compliant linear-motion mechanism based on parasitic motion compensation. *Mechanism and Machine Theory* 50: 15-28.

Zhu J, Cai S and Suo Z (2010) Resonant behavior of a membrane of a dielectric elastomer. *International Journal of Solids and Structures* 47(24): 3254-3262.